\newif\ifJOURNAL
\JOURNALfalse
\newif\ifCONF
\CONFfalse
\newif\ifarXiv
\arXivfalse
\newif\ifWP
\WPfalse
\newif\ifFULL
\FULLfalse
\newif\ifLATIN
\LATINfalse

\arXivtrue

\ifarXiv\LATINtrue\fi	

\newif\ifnotCONF	
\notCONFtrue
\ifCONF\notCONFfalse\fi

\newif\ifnotarXiv	
\notarXivtrue
\ifarXiv\notarXivfalse\fi

\newif\ifTR		
\TRfalse
\ifarXiv\TRtrue\fi
\ifWP\TRtrue\fi
\newif\ifnotTR
\notTRtrue
\ifarXiv\notTRfalse\fi
\ifWP\notTRfalse\fi

\newif\ifnotLATIN	
\notLATINtrue
\ifLATIN\notLATINfalse\fi

\ifCONF

  \newcommand{\DFIII}{vovk:2005ALT-DF03}
  
  \newcommand{\DFV}{vovk:2006China}
  \newcommand{\DFVI}{vovk:2006COLT-DF06}
  \newcommand{\DFVII}{vovk:2006COLT-DF07}
  \newcommand{\DFVIII}{DF08arXiv}
  \newcommand{\DFIX}{DF09arXiv}
  \newcommand{\DFX}{DF10arXiv}
\fi
\ifarXiv

  \newcommand{\DFIII}{DF03arXiv}
  
  \newcommand{\DFV}{DF05arXiv}
  \newcommand{\DFVI}{DF06arXiv-local}
  \newcommand{\DFVII}{DF07arXiv-local}
  \newcommand{\DFVIII}{DF08arXiv-local}
  \newcommand{\DFIX}{DF09arXiv-local}
  \newcommand{\DFX}{DF10arXiv-local}
\fi
\ifWP

  \newcommand{\DFIII}{GTP13}
  
  \newcommand{\DFV}{GTP11}
  \newcommand{\DFVI}{GTP16}
  \newcommand{\DFVII}{GTP17}
  \newcommand{\DFVIII}{DF08arXiv}
  \newcommand{\DFIX}{DF09arXiv}
  \newcommand{\DFX}{DF10arXiv}
\fi
\ifFULL

  \newcommand{\DFIII}{DF03arXiv}
  
  \newcommand{\DFV}{DF05arXiv}
  \newcommand{\DFVI}{DF06arXiv-local}
  \newcommand{\DFVII}{DF07arXiv-local}
  \newcommand{\DFVIII}{DF08arXiv-local}
  \newcommand{\DFIX}{DF09arXiv-local}
  \newcommand{\DFX}{DF10arXiv-local}
\fi

\ifnotLATIN
  \newcommand{\Bary}{bary:1964}
  \newcommand{\KolmogorovTikhomirov}{kolmogorov/tikhomirov:1959}
  \newcommand{\MarkushevichII}{markushevich:1968}
  
  \newcommand{\Vitushkin}{vitushkin:1959}
\fi
\ifLATIN
  \newcommand{\Bary}{bary:1964latin}
  \newcommand{\KolmogorovTikhomirov}{kolmogorov/tikhomirov:1959latin}
  \newcommand{\MarkushevichII}{markushevich:1968latin}
  
  \newcommand{\Vitushkin}{vitushkin:1959latin}
\fi

\ifCONF
\documentclass{llncs}
\usepackage{amsmath,amsfonts,amssymb,latexsym,graphicx,mathrsfs}
\newcommand{\Extra}[1]{}
\fi

\ifarXiv
\documentclass{article}
\usepackage{amsmath,amsfonts,amssymb,latexsym,graphicx,mathrsfs}
\newcommand{\Extra}[1]{}
\fi

\ifWP
\documentclass[toc]{gtarticle}
\usepackage{amsmath,amsfonts,amssymb,latexsym,epsfig,graphicx,mathrsfs}
\renewcommand{\Extra}[1]{#1}
\fi

\ifFULL
\documentclass{article}
\usepackage{amsmath,amsfonts,amssymb,latexsym,color,graphicx,mathrsfs}
\newcommand{\Extra}[1]{\red{#1}}
\newcommand{\red}[1]{\textcolor{red}{#1}}

\newcommand{\bluebegin}{\begingroup\color{blue}}
\newcommand{\blueend}{\endgroup}

\fi

\ifnotLATIN
\usepackage{epigraph}
\fi

\allowdisplaybreaks[3]

\ifnotLATIN
\input{OT2enc.def}
\newenvironment{cyr}
{\fontencoding{OT2}\fontfamily{wncyr}\fontseries{m}\fontshape{n}\selectfont}
{\fontencoding{OT1}\fontfamily{tir}\selectfont}
\fi

\newcommand{\Vladimir}{Vladimir}
\newcommand{\DOT}{.}
\newcommand{\zzrelax}[1]{}

\newcommand{\st}{\mathrel{\!|\!}}
\newcommand{\stbegin}{\left|\;}
\newcommand{\stend}{\right.}
\newcommand{\stbig}{\;\bigl|\;}
\newcommand{\stBig}{\;\Bigl|\;}

\newcommand{\givn}{\mathrel{|}}

\newcommand{\dd}{\mathrm{d}}
\renewcommand{\Im}{\mathop{\mathrm{Im}}}
\newcommand{\risk}{\mathop{\mathrm{risk}}\nolimits}

\newcommand{\AAA}{\mathcal{A}}		
\newcommand{\CCC}{\mathscr{C}}		
\newcommand{\EEE}{\mathcal{E}}		
\newcommand{\FFF}{\mathcal{F}}		
\newcommand{\HHH}{\mathcal{H}}		
\newcommand{\KKK}{\mathbf{K}}		
\newcommand{\ccc}{\mathbf{c}}		

\ifnotCONF
  \newcommand{\bbbp}{\mathbb{P}}		
\fi
\ifCONF
  \renewcommand{\bbbp}{\mathbb{P}}		
\fi
\newcommand{\Prob}{\mathop{\bbbp}\nolimits}
\newcommand{\bbbe}{\mathbb{E}}		
\newcommand{\Expect}{\mathop{\bbbe}\nolimits}

\ifnotCONF
\newcommand{\bbbr}{\mathbb{R}}		
\newcommand{\bbbc}{\mathbb{C}}		
\newcommand{\bbbd}{\mathbb{D}}		
\newtheorem{lemma}{Lemma}
\newtheorem{proposition}{Proposition}
\newtheorem{corollary}{Corollary}
\newtheorem{theorem}{Theorem}
\newenvironment{proof}
  {\trivlist\item[\hskip\labelsep\textbf{Proof}]}
  {\endtrivlist}
\fi

\newenvironment{Proof}[1]
  {\trivlist\item[\hskip\labelsep\textbf{Proof #1\;}]}
  {\endtrivlist}
\newcommand{\boxforqed}{\rule{.3em}{1.5ex}}
\newcommand{\qedtext}{\unskip\nobreak\hfil
  \penalty50\hskip1em\null\nobreak\hfil\boxforqed
  \parfillskip=0pt\finalhyphendemerits=0\endgraf}
\newcommand{\qedmath}{\tag*{\boxforqed}}
\newenvironment{remark*}
  {\trivlist\item[\hskip\labelsep{\bfseries Remark}]\relax}
  {\endtrivlist}
\newenvironment{problem*}
  {\trivlist\item[\hskip\labelsep{\bfseries Problem}]\relax}
  {\endtrivlist}

\newlength{\IndentI}
\newlength{\IndentII}
\newlength{\IndentIII}
\newlength{\IndentIV}
\newlength{\WidthI}
\newlength{\WidthII}
\newlength{\WidthIII}
\newlength{\WidthIV}
\ifCONF
\title{Metric entropy in competitive on-line prediction}
\author{Vladimir Vovk}
\institute{Computer Learning Research Centre,
  Department of Computer Science\\
  Royal Holloway, University of London,
  Egham, Surrey TW20 0EX, UK\\
  \email{vovk@cs.rhul.ac.uk}}
\fi

\ifarXiv
\title{Metric entropy in competitive on-line prediction}
\author{Vladimir Vovk\\
\texttt{vovk{\rm@}cs.rhul.ac.uk}\\
\texttt{http://vovk.net}}
\fi

\ifWP
\title{Metric entropy in competitive on-line prediction}
\author{Vladimir Vovk}

\twodatestrue

\fi

\ifFULL
\title{Metric entropy in competitive on-line prediction}
\author{Vladimir Vovk\\
\texttt{vovk{\rm@}cs.rhul.ac.uk}\\
\texttt{http://vovk.net}}
\fi

\begin{document}
\maketitle
\begin{abstract}
  Competitive on-line prediction
  (also known as universal prediction of individual sequences)
  is a strand of learning theory avoiding making any stochastic assumptions
  about the way the observations are generated.
  The predictor's goal is to compete with a benchmark class of prediction rules,
  which is often a proper Banach function space.
  Metric entropy provides a unifying framework for competitive on-line prediction:
  the numerous known upper bounds on the metric entropy of various compact sets
  in function spaces readily imply bounds on the performance
  of on-line prediction strategies.
  This paper discusses strengths and limitations of the direct approach
  to competitive on-line prediction via metric entropy,
  including comparisons to other approaches.
\end{abstract}

\ifnotLATIN
\setlength{\epigraphwidth}{0.55\textwidth}
\epigraph{\begin{cyr}Voob{shch}e mne predstavl{ya}et{}s{ya} va{zh}no\cyrishrt\
  zada{ch}a osvobo{zh}deni{ya} vs{yu}du,
  gde \cyrerev{}to vozmo{zh}no,
  ot izli{sh}nih vero{ya}tnostnyh dopu{shch}eni\cyrishrt.\end{cyr}}
{Andrei Kolmogorov, 1987}
\fi

\section{Introduction}
\label{sec:introduction}

A typical result of competitive on-line prediction
says that, for a given benchmark class of prediction strategies,
there is a prediction strategy that performs almost as well as the best prediction strategies
in the benchmark class.
For simplicity,
in this paper the performance of a prediction strategy will be measured
by the cumulative squared distance between its predictions and the true observations,
assumed to be real (occasionally complex) numbers.
Different methods of competitive on-line predictions
(such as Gradient Descent,
following the perturbed leader,
strong and weak aggregating algorithms,
defensive forecasting, etc.)\ tend to have their narrow ``area of expertise'':
each works well for benchmark classes of a specific ``size''
but is not readily applicable to classes of a different size.

In this paper we will apply a simple general method
based on metric entropy
to benchmark classes of a wide range of sizes.
Typically,
this method does not give optimal results,
but its results are often not much worse than those given by specialized methods,
especially for benchmark classes that are not too massive.
Since the method is almost universally applicable,
it sheds new light on the known results.

Another disadvantage of the metric entropy method
is that it is not clear how to implement it efficiently,
whereas many other methods are computationally very efficient.
Therefore,
the results obtained by this method are only a first step,
and we should be looking for other prediction strategies,
both computationally more efficient and having better performance guarantees.

We start, in \S\ref{sec:asymptotic},
by stating a simple asymptotic result
about the existence of a universal prediction strategy
for the class of continuous prediction rules.
The performance of the universal strategy is in the long run
as good as the performance of any continuous prediction rule,
but we do not attempt to estimate the rate at which the former approaches the latter.
This is the topic of the following section,
\S\ref{sec:entropy},
where we establish general results
about performance guarantees based on metric entropy.
For example, in the simplest case where the benchmark class $\FFF$
is a compact set,
the performance guarantees become weaker
as the metric entropy of $\FFF$ becomes larger.

The core of the paper is organized according to the types of metric compacts
pointed out by Kolmogorov and Tikhomirov in \cite{\KolmogorovTikhomirov} (\S3).
Type I compacts have metric entropy of order $\log\frac{1}{\epsilon}$;
this case corresponds to the finite-dimensional benchmark classes
and is treated in \S\ref{sec:finite-dimensional}.
Type II, with the typical order $\log^M\frac{1}{\epsilon}$,
contains various classes of analytic functions
and is dealt with in \S\ref{sec:analytic}.
The key \S\ref{sec:Sobolev} deals with perhaps the most important case
of order $\left(\frac{1}{\epsilon}\right)^{\gamma}$;
this includes, e.g., Besov classes.
The classes of type IV, considered in \S\ref{sec:image}, have metric entropy
that grows even faster.

In \S\S\ref{sec:finite-dimensional}--\ref{sec:image} the benchmark class
is always given.
In \S\ref{sec:super} we ask the question of how
prediction strategies competitive against various benchmark classes
compare to each other.
The previous section, \S\ref{sec:norm}, prepares the ground for this.
The concluding section, \S\ref{sec:conclusion},
lists several directions of further research.

There is no real novelty in this paper;
I just apply known results about metric entropy to competitive on-line prediction.
I hope it will be useful as a survey.

\section{Simple asymptotic result}
\label{sec:asymptotic}

Throughout the paper we will be interested in the following prediction protocol
(or its modifications):

\bigskip

\noindent
\textsc{On-line regression protocol}\nopagebreak

\parshape=5
\IndentI  \WidthI
\IndentII \WidthII
\IndentII \WidthII
\IndentII \WidthII
\IndentI  \WidthI
\noindent
FOR $n=1,2,\dots$:\\
  Reality announces $x_n\in\mathbf{X}$.\\
  Predictor announces $\mu_n\in\bbbr$.\\
  Reality announces $y_n\in[-Y,Y]$.\\
END FOR.

\bigskip

\noindent
At the beginning of each round $n$ Predictor is given some \emph{signal} $x_n$
that might be helpful in predicting the following \emph{observation} $y_n$,
after which he announces his \emph{prediction} $\mu_n$.
The signal is taken from the \emph{signal space} $\mathbf{X}$,
the observations are real numbers
known to belong to a fixed interval $[-Y,Y]$, $Y>0$,
and the predictions are any real numbers
(later this will also be extended to complex numbers).
The error of prediction is always measured by the quadratic loss function,
so the loss suffered by Predictor on round $n$
is $(y_n-\mu_n)^2$.
It is clear that it never makes sense for Predictor
to choose predictions outside $[-Y,Y]$,
but the freedom to go outside $[-Y,Y]$ might be useful
when the benchmark class is not closed under truncation.

\begin{remark*}
  Competitive on-line prediction uses a wide range of loss functions
  $\lambda(y_n,\mu_n)$.
  The quadratic loss function
  $\lambda(y_n,\mu_n):=(y_n-\mu_n)^2$
  belongs to the class of ``mixable'' loss functions,
  which are strictly convex in the prediction $\mu_n$ in a fairly strong sense.
  Such loss functions allow the strongest performance guarantees
  (using, e.g., the ``aggregating algorithm'' of \cite{vovk:2001competitive},
  which we will call the strong aggregating algorithm).
  If the loss function is convex but not strictly convex in the prediction,
  the performance guarantees somewhat weaken
  (and can be obtained using, e.g., the weak aggregating algorithm of \cite{kalnishkan/vyugin:2005};
  for a review of earlier methods, see \cite{cesabianchi/lugosi:2006}).
  When the loss function is not convex in the prediction,
  it can be ``convexified'' by using randomization
  (\cite{cesabianchi/lugosi:2006}, Chapter 4).
\end{remark*}

A \emph{prediction rule} is a function $F:\mathbf{X}\to\bbbr$.
Intuitively,
$F$ plays the role of the strategy for Predictor that recommends prediction $F(x_n)$
after observing signal $x_n\in\mathbf{X}$;
such strategies are called Markov prediction strategies in \cite{\DFIX}.
We will be interested in benchmark classes
consisting of only Markov prediction strategies;
in practice, this is not as serious a restriction as it might appear:
it is usually up to us what we want to include in the signal space $\mathbf{X}$,
and we can always extend $\mathbf{X}$ by including, e.g.,
some of the previous observations and signals.

Our first result states the existence of a strategy for Predictor
that asymptotically dominates every continuous prediction rule
(for much stronger asymptotic results,
see \cite{\DFVII,\DFVIII,\DFIX,\DFX}).
\begin{theorem}\label{thm:asymptotic}
  Let $\mathbf{X}$ be a metric compact.
  There exists a strategy for Predictor that guarantees
  \begin{equation}\label{eq:asymptotic}
    \limsup_{N\to\infty}
    \left(
      \frac1N
      \sum_{n=1}^N
      \left(
        y_n-\mu_n
      \right)^2
      -
      \frac1N
      \sum_{n=1}^N
      \left(
        y_n-F(x_n)
      \right)^2
    \right)
    \le
    0
  \end{equation}
  for each continuous prediction rule $F$.
\end{theorem}
Any strategy for Predictor that guarantees (\ref{eq:asymptotic})
for each continuous $F:\mathbf{X}\to\bbbr$
will be said to be \emph{universal}
(or, more fully, \emph{universal for $C(\mathbf{X})$}).
Theorem \ref{thm:asymptotic},
asserting the existence of universal prediction strategies,
will be proved at the end of this section.

\subsection*{Aggregating algorithm}

This subsection will introduce the main technical tool used in this paper,
an aggregating algorithm
(in fact intermediate between the strong aggregating algorithm of \cite{vovk:2001competitive}
and the weak aggregating algorithm of \cite{kalnishkan/vyugin:2005}).
For future use in \S\ref{sec:analytic},
we will allow the observations to belong to the Euclidean space $\bbbr^m$
(in fact, we will only be interested in the cases $m=1$ and $m=2$).
Correspondingly,
we allow predictions in $\bbbr^m$
and extend the notion of prediction rule allowing values in $\bbbr^m$.

The $l_2$ norm in $\bbbr^m$ will be denoted $\left\|\cdot\right\|_2$
or simply $\left\|\cdot\right\|$;
in later sections we will also use $l_p$ norms $\left\|\cdot\right\|_p$ for $p\ne2$.
Reality is constrained to producing observations in the ball $YU_m$ in $\bbbr^m$
with radius $Y$ and centred at $0$;
$U_V$ is our general notation for the closed unit ball
$\{v\in V\st\left\|v\right\|\le1\}$
centred at $0$ in a Banach space $V$,
and we abbreviate $U_m:=U_{\bbbr^m}$.

\begin{lemma}\label{lem:AA}
  Let $F_1,F_2,\ldots$ be a sequence of $\bbbr^m$-valued prediction rules
  assigned positive weights $w_1,w_2,\ldots$ summing to $1$.
  There is a strategy for Predictor
  producing $\mu_n\in Y U_m$ that are guaranteed to satisfy,
  for all $N=1,2,\ldots$ and all $i=1,2,\ldots$,
  \begin{equation}\label{eq:AA}
    \sum_{n=1}^N
    \left\|
      y_n-\mu_n
    \right\|^2
    \le
    \sum_{n=1}^N
    \left\|
      y_n-F_i(x_n)
    \right\|^2
    +
    8Y^2
    \ln\frac{1}{w_i}.
  \end{equation}
\end{lemma}
For $m=1$ the constant $8Y^2$ in (\ref{eq:AA})
can be improved to $2Y^2$
(cf.\ \cite{vovk:2001competitive},
Lemma 2 and the line above Remark 3),
and it is likely that this is also true in general.
In this paper we, however, do not care about multiplicative constants
(and usually even do not give them explicitly in the statements of our results;
the reader can always extract them from the proofs).

Inequality (\ref{eq:AA}) says that Predictor's total loss
does not exceed the total loss suffered by an alternative prediction strategy
plus a \emph{regret term}
($
  8Y^2
  \ln\frac{1}{w_i}
$
in the case of (\ref{eq:AA}));
we will encounter many such inequalities in the rest of this paper.

\begin{Proof}{of Lemma \ref{lem:AA}}
  Let $\eta:=\frac{1}{8Y^2}$, $\beta:=e^{-\eta}$,
  and $P_0$ be the probability measure on $\{1,2,\ldots\}$
  assigning weight $w_i$ to each $i=1,2,\ldots$\,.
  Lemma 1 and Remark 3 of \cite{vovk:2001competitive} imply that it suffices
  to show that the function $\beta^{\left\|y-\mu\right\|^2}$
  is concave in $\mu\in YU_m$ for each fixed $y\in YU_m$
  (this idea goes back to Kivinen and Warmuth \cite{kivinen/warmuth:1999}).
  Furthermore,
  it suffices to show that the function
  \begin{equation*}
    \beta^{\left\|a+bt\right\|^2}
    =
    e^{
      -\eta
      \left(
        \left\|a\right\|^2
        +
        2 \langle a,b\rangle t
        +
        \left\|b\right\|^2 t^2
      \right)
    }
  \end{equation*}
  is convex in $t\in[0,1]$ for any $a$ and $b$
  such that $a$ and $a+b$ belong to the ball $2YU_m$ of radius $2Y$ centred at $0$.
  Taking the second derivative,
  we can see that we need to show
  \begin{equation*}
    2\eta
    \bigl(
      \langle a,b\rangle
      +
      \left\|b\right\|^2 t
    \bigr)^2
    \le
    \left\|b\right\|^2.
  \end{equation*}
  By the convexity of the function $(\cdot)^2$,
  it suffices to establish the last inequality for $t=0$,
  \begin{equation}\label{eq:concave-1}
    2\eta
    \bigl(
      \langle a,b\rangle
    \bigr)^2
    \le
    \left\|b\right\|^2,
  \end{equation}
  and $t=1$,
  \begin{equation}\label{eq:concave-2}
    2\eta
    \bigl(
      \langle a,b\rangle
      +
      \left\|b\right\|^2
    \bigr)^2
    \le
    \left\|b\right\|^2.
  \end{equation}
  Inequality (\ref{eq:concave-1}) follows, for $\eta\le\frac{1}{8Y^2}$, from
  \begin{equation*}
    2\eta
    \bigl(
      \langle a,b\rangle
    \bigr)^2
    \le
    2\eta
    \left\|a\right\|^2
    \left\|b\right\|^2
    \le
    8\eta
    Y^2
    \left\|b\right\|^2
    \le
    \left\|b\right\|^2.
  \end{equation*}
  In the case of (\ref{eq:concave-2}),
  it is clear that we can replace $a$ by its projection onto the direction of $b$
  and so assume $a=\lambda b$ for some $\lambda\in\bbbr$.
  Therefore, (\ref{eq:concave-2}) becomes
  \begin{equation*}
    2\eta
    (1+\lambda)^2
    \left\|b\right\|^4
    \le
    \left\|b\right\|^2.
  \end{equation*}
  The last inequality,
  equivalent to
  $
    2\eta
    \left\|(1+\lambda)b\right\|^2
    \le
    1
  $,
  immediately follows from the fact that $(1+\lambda)b=a+b$ belongs to $2YU_{m}$.
  \qedtext
\end{Proof}

The proof of Lemma \ref{lem:AA} exhibits an explicit strategy for Predictor
guaranteeing (\ref{eq:AA});
we will refer to this strategy as the \emph{AA mixture} of $F_1,F_2,\ldots$
(with weights $w_1,w_2,\ldots$).

\subsection*{Proof of Theorem \ref{thm:asymptotic}}

Theorem \ref{thm:asymptotic} follows immediately from the separability
of the function space $C(\mathbf{X})$ of continuous real-valued functions on $\mathbf{X}$
(\cite{engelking:1989}, Corollary 4.2.18).
Indeed, we can choose a dense sequence $F_1,F_2,\ldots$ of prediction rules
in $C(\mathbf{X})$
and take any positive weights $w_i$ summing to $1$.
Let $F$ be any continuous prediction rule;
without loss of generality,
$F:\mathbf{X}\to[-Y,Y]$.
For any $\epsilon>0$,
the AA mixture clipped to $[-Y,Y]$ will satisfy
\begin{multline*}
  \limsup_{N\to\infty}
  \left(
    \frac1N
    \sum_{n=1}^N
    \left(
      y_n-\mu_n
    \right)^2
    -
    \frac1N
    \sum_{n=1}^N
    \left(
      y_n-F(x_n)
    \right)^2
  \right)\\
  \le
  \limsup_{N\to\infty}
  \left(
    \frac1N
    \sum_{n=1}^N
    \left(
      y_n-\mu_n
    \right)^2
    -
    \frac1N
    \sum_{n=1}^N
    \left(
      y_n-F_i(x_n)
    \right)^2
  \right)
  +4Y\epsilon\\
  \le
  \limsup_{N\to\infty}
  \frac{8Y^2\ln\frac{1}{w_i}}{N}
  +
  4Y\epsilon
  \le
  4Y\epsilon,
\end{multline*}
where $i$ is such that $F_i$ is $\epsilon$-close to $F$ in $C(\mathbf{X})$.
Since this holds for any $\epsilon>0$,
the proof is complete.

\section{Performance guarantees based on metric entropy: general results}
\label{sec:entropy}

In the rest of the paper we will be assuming,
without loss of generality,
that $Y=1$.
To recover the case of a general $Y>0$,
the universal constants $C$ in Theorems \ref{thm:compact}--\ref{thm:continuous}
(and their corollaries) below
should be replaced by $CY^2$
and all the norms should be divided by $Y$.
The constants $C$ in those theorems are not too large
(of order $10$ according to the proofs given,
but no effort has been made to optimize them).

We will consider three types of non-asymptotic versions
of Theorem \ref{thm:asymptotic},
corresponding to Theorems \ref{thm:compact}--\ref{thm:continuous}
of this section.
In the first type the benchmark class $\FFF$ is a metric compact,
and we can guarantee that Predictor's loss
over the first $N$ observations does not exceed the loss suffered
by the best prediction rule in the benchmark class plus a regret term of $o(N)$,
the rate of growth of the regret term depending on the metric entropy of $\FFF$.
In the second type $\FFF$ is a Banach function space on $\mathbf{X}$
whose unit ball $U_{\FFF}$ is a compact (in metric $C(\mathbf{X})$)
subset of $C(\mathbf{X})$.
In this case it is impossible to have the same performance guarantees;
Predictor will need a start (given in terms of their norm in the Banach space)
on remote prediction rules.
Results of this type can be easily obtained from results of the first type.
In the third type the benchmark class consists of all continuous prediction rules;
such results can be obtained from results of the second type
for ``universal'' Banach spaces,
i.e., Banach spaces that are dense subsets of $C(\mathbf{X})$.

\subsection*{Compact benchmark classes}

Let $A$ be a compact metric space.
The \emph{metric entropy} $\HHH_{\epsilon}(A)$,
$\epsilon>0$, is defined to be the binary logarithm $\log N$
of the minimum number of elements
$F_1,\ldots,F_N\in A$
that form an $\epsilon$-net for $A$
(in the sense that for each $F\in A$ there exists $i=1,\ldots,N$
such that $F$ and $F_i$ are $\epsilon$-close in $A$).
The requirement of compactness of $A$ ensures that $\HHH_{\epsilon}(A)$
is finite for each $\epsilon>0$.

\begin{remark*}
  There are four main variations on the notion of metric entropy
  as defined in \cite{\KolmogorovTikhomirov};
  our definition corresponds to Kolmogorov and Tikhomirov's
  relative $\epsilon$-entropy $\HHH_{\epsilon}^A(A)$.
  In general, relative $\epsilon$-entropy $\HHH_{\epsilon}^R(A)$
  can be defined for any metric space $R$ containing $A$ as a subspace
  (in our applications we would take $R:=C(\mathbf{X})$).
  The other two variations are the absolute $\epsilon$-entropy $\HHH^{\textrm{abs}}_{\epsilon}(A)$
  (denoted simply $\HHH_{\epsilon}(A)$ by Kolmogorov and Tikhomirov;
  it was introduced by Pontryagin and Shnirel'man \cite{pontryagin/shnirelman:1932}
  in 1932, without taking the binary logarithm
  and using $\epsilon$ in place of Kolmogorov and Tikhomirov's $2\epsilon$)
  and the $\epsilon$-capacity $\EEE_{\epsilon}(A)$.
  All four notions were studied by Kolmogorov, his students
  (Vitushkin, Erokhin, Tikhomirov, Arnol'd),
  and Babenko in the 1950s,
  and their results are summarized in \cite{\KolmogorovTikhomirov}.
  It is always true that, in our notation,
  \begin{equation}\label{eq:chain}
    \EEE_{2\epsilon}(A)
    \le
    \HHH^{\textrm{abs}}_{\epsilon}(A)
    \le
    \HHH^{R}_{\epsilon}(A)
    \le
    \HHH_{\epsilon}(A)
    \le
    \EEE_{\epsilon}(A)
  \end{equation}
  (\cite{\KolmogorovTikhomirov}, Theorem IV).
  All results in \cite{\KolmogorovTikhomirov}
  can be applied to all elements of the chain (\ref{eq:chain}),
  and in principle we can use any of the four notions;
  our choice of $\HHH_{\epsilon}(A)=\HHH^A_{\epsilon}(A)$ is closest
  to the notion of entropy numbers
  popular in the recent literature
  (such as \cite{carl/stephani:1990}).
\end{remark*}

\begin{theorem}\label{thm:compact}
  Suppose $\FFF$ is a compact set in $C(\mathbf{X})$.
  There exists a strategy for Predictor
  that produces $\mu_n$ with $\left|\mu_n\right|\le1$
  and guarantees, for all $N=1,2,\ldots$ and all $F\in\FFF$,
  \begin{equation}\label{eq:compact}
    \sum_{n=1}^N
    \left(
      y_n-\mu_n
    \right)^2
    \le
    \sum_{n=1}^N
    \left(
      y_n-F(x_n)
    \right)^2
    +
    C
    \inf_{\epsilon\in\left(0,1/2\right]}
    \left(
      \HHH_{\epsilon}(\FFF)
      +
      \log\log\frac{1}{\epsilon}
      +
      \epsilon N
      +
      1
    \right),
  \end{equation}
  where $C$ is a universal constant.
\end{theorem}
\begin{proof}
  Without loss of generality we can only consider $\epsilon$
  of the form $2^{-i}$, $i=1,2,\ldots$,
  in (\ref{eq:compact}).
  Let us fix, for each $i$,
  a $2^{-i}$-net $\FFF_i$ for $\FFF$ of size $2^{\HHH_{2^{-i}}(\FFF)}$;
  to each element of $\FFF_i$ we assign weight $\frac{6}{\pi^2}i^{-2}2^{-\HHH_{2^{-i}}(\FFF)}$,
  so that the weights sum to $1$.
  Our goal (\ref{eq:compact}) will be achieved
  if we establish,
  for each $i=1,2,\ldots$,
  \begin{equation}\label{eq:goal}
    \sum_{n=1}^N
    \left(
      y_n-\mu_n
    \right)^2
    \le
    \sum_{n=1}^N
    \left(
      y_n-F(x_n)
    \right)^2
    +
    C
    \inf_{i=1,2,\ldots}
    \Bigl(
      \HHH_{2^{-i}}(\FFF)
      +
      \log i
      +
      2^{-i} N
      +
      1
    \Bigr)
  \end{equation}
  (we let $C$ stand for different constants in different formulas).

  Without loss of generality it will be assumed that $F$
  and all functions in $\FFF_i$, $i=1,2,\ldots$,
  take values in $[-1,1]$.
  Fix an $i$.
  Let $F^*\in\FFF_i$ be $2^{-i}$-close to $F$ in $C(\mathbf{X})$.
  Lemma \ref{lem:AA} gives a prediction strategy satisfying
  \begin{multline*}
    \sum_{n=1}^N
    \left(
      y_n-\mu_n
    \right)^2
    \le
    \sum_{n=1}^N
    \left(
      y_n-F^*(x_n)
    \right)^2
    +
    8
    \ln
    \left(
      \frac{\pi^2}{6}
      i^{2}2^{\HHH_{2^{-i}}(\FFF)}
    \right)\\
    \le
    \sum_{n=1}^N
    \left(
      y_n-F(x_n)
    \right)^2
    +
    8
    \ln
    \left(
      \frac{\pi^2}{6}
      i^{2}2^{\HHH_{2^{-i}}(\FFF)}
    \right)
    +
    4\left(2^{-i}N\right)\\
    \le
    \sum_{n=1}^N
    \left(
      y_n-F(x_n)
    \right)^2
    +
    C
    \left(
      1
      +
      \log i
      +
      \HHH_{2^{-i}}(\FFF)
      +
      2^{-i}N
    \right),
  \end{multline*}
  which coincides with (\ref{eq:goal}).
  \qedtext
\end{proof}

\subsection*{Banach function spaces as benchmark classes}

Let $\FFF$ be a linear subspace of $C(\mathbf{X})$
equipped with a norm making it into a Banach space.
We will be interested in the case where $\FFF$
is \emph{compactly embedded} into $C(\mathbf{X})$,
in the sense that the unit ball
\begin{equation*}
  U_{\FFF}
  :=
  \left\{
    F\in\FFF
    \st
    \left\|F\right\|_{\FFF}\le1
  \right\}
\end{equation*}
is a compact subset of $C(\mathbf{X})$.
(The Arzel\`a--Ascoli theorem, \cite{dudley:2002}, 2.4.7,
shows that all such $\FFF$ are Banach function spaces with finite embedding constant,
as defined in \cite{\DFVI} and below;
in particular,
they are proper Banach functional spaces.)

\begin{theorem}\label{thm:Banach}
  Let $\FFF$ be a Banach space compactly embedded in $C(\mathbf{X})$.
  There exists a strategy for Predictor
  that produces $\mu_n$ with $\left|\mu_n\right|\le1$
  and guarantees, for all $N=1,2,\ldots$ and all $F\in\FFF$,
  \begin{multline}\label{eq:Banach}
    \sum_{n=1}^N
    \left(
      y_n-\mu_n
    \right)^2
    \le
    \sum_{n=1}^N
    \left(
      y_n-F(x_n)
    \right)^2\\
    +
    C
    \inf_{\epsilon\in\left(0,1/2\right]}
    \left(
      \HHH_{\epsilon/\phi}(U_{\FFF})
      +
      \log\log\frac{1}{\epsilon}
      +
      \log\log\phi
      +
      \epsilon N
      +
      1
    \right),
  \end{multline}
  where $C$ is a universal constant and
  \begin{equation}\label{eq:phi}
    \phi
    :=
    2\max
    \bigl(
      1,
      \left\|F\right\|_{\FFF}
    \bigr).
  \end{equation}
\end{theorem}
\begin{proof}
  Notice that $\HHH_{\epsilon}(2^i U_{\FFF}) = \HHH_{2^{-i}\epsilon}(U_{\FFF})$,
  $i=1,2,\ldots$\,.
  Applying (\ref{eq:compact}) to $\FFF:=2^i U_{\FFF}$,
  we obtain
  \begin{equation}\label{eq:component}
    \sum_{n=1}^N
    \left(
      y_n-\mu_n
    \right)^2
    \le
    \sum_{n=1}^N
    \left(
      y_n-F(x_n)
    \right)^2
    +
    C
    \left(
      \HHH_{2^{-i}\epsilon}(U_{\FFF})
      +
      \log\log\frac{1}{\epsilon}
      +
      \epsilon N
      +
      1
    \right)
  \end{equation}
  for any $\epsilon\in\left(0,1/2\right]$;
  we will assign weight $\frac{6}{\pi^2}i^{-2}$
  to the corresponding prediction strategy.
  AA mixing the prediction strategies achieving (\ref{eq:component}) for $i=1,2,\ldots$,
  (it is clear that Lemma \ref{lem:AA} is applicable to any prediction strategies,
  not only prediction rules),
  we obtain a strategy achieving
  \begin{multline}\label{eq:mixed}
    \sum_{n=1}^N
    \left(
      y_n-\mu_n
    \right)^2
    \le
    \sum_{n=1}^N
    \left(
      y_n-F(x_n)
    \right)^2
    +
    C
    \left(
      \HHH_{2^{-i}\epsilon}(U_{\FFF})
      +
      \log\log\frac{1}{\epsilon}
      +
      \epsilon N
      +
      1
    \right)\\
    +
    8
    \ln
    \left(
      \frac{\pi^2}{6}
      i^2
    \right)
  \end{multline}
  for all $i=1,2,\ldots$ and all $F\in2^iU_{\FFF}$.
  For each $F\in\FFF$ we can set
  \begin{equation*}
    i
    :=
    \max
    \left(
      1,
      \left\lceil
        \log
        \left\|
          F
        \right\|_{\FFF}
      \right\rceil
    \right)
  \end{equation*}
  to obtain $2^{i}\le\phi\le2^{i+1}$ and so, from (\ref{eq:mixed}),
  \begin{multline*}
    \sum_{n=1}^N
    \left(
      y_n-\mu_n
    \right)^2
    \le
    \sum_{n=1}^N
    \left(
      y_n-F(x_n)
    \right)^2
    +
    C
    \left(
      \HHH_{\epsilon/\phi}(U_{\FFF})
      +
      \log\log\frac{1}{\epsilon}
      +
      \epsilon N
      +
      1
    \right)\\
    +
    8
    \ln
    \left(
      \frac{\pi^2}{6}
      \log^2\phi
    \right).
  \end{multline*}
  The last inequality can be written as (\ref{eq:Banach}).
  \qedtext
\end{proof}

\subsection*{Competing with the continuous prediction rules}

Let $\FFF\subseteq C(\mathbf{X})$ be a Banach function space
(no connection between the norms in $\FFF$ and $C(\mathbf{X})$ is assumed)
which is dense in $C(\mathbf{X})$
(in the $C(\mathbf{X})$ metric, of course);
in this case we will say that $\FFF$ is \emph{densely embedded} in $C(\mathbf{X})$.
The \emph{approachability} of $F\in C(\mathbf{X})$ by $\FFF$ is defined as the function
\begin{equation}\label{eq:approachability}
  \AAA_{\epsilon}^{\FFF}(F)
  :=
  \inf
  \left\{
    \left\|
      F^*
    \right\|_{\FFF}
    \stbegin
    \left\|
      F-F^*
    \right\|_{C(\mathbf{X})}
    \le
    \epsilon
    \stend
  \right\},
  \quad
  \epsilon>0,
\end{equation}
which is finite under our assumption of density.

\begin{remark*}
  The \emph{Gagliardo set} of a function $F\in C(\mathbf{X})$
  can be defined as
  \begin{multline}\label{eq:Gagliardo}
    \Gamma(F)
    :=
    \Bigl\{
      (t_0,t_1)\in\bbbr^2
      \stBig
      \exists F_0\in C(\mathbf{X}),F_1\in\FFF:
      F_0+F_1=F,\\
      \left\|
        F_0
      \right\|_{C(\mathbf{X})}
      \le
      t_0,
      \left\|
        F_1
      \right\|_{\FFF}
      \le
      t_1
    \Bigr\}.
  \end{multline}
  (See \cite{bergh/lofstrom:1976}, \S3.1, for the general definition.)
  The graph of the function $\epsilon\mapsto\AAA_{\epsilon}^{\FFF}(F)$
  is essentially the boundary of $\Gamma(F)$.
  A third way of talking about the Gagliardo set
  is in terms of the norm
  \begin{equation}\label{eq:K}
    K(t,F)
    :=
    \inf_{F_0\in\FFF,F_1\in C(\mathbf{X}):F=F_0+F_1}
    \left(
      \left\|
        F_0
      \right\|_{C(\mathbf{X})}
      +
      t
      \left\|
        F_1
      \right\|_{\FFF}
    \right),
  \end{equation}
  where $t$ ranges over the positive numbers.
  (See \cite{bergh/lofstrom:1976}, \S3.1, or \cite{adams/fournier:2003}, 7.8,
  for further details.)
\end{remark*}

\begin{theorem}\label{thm:continuous}
  Let $\FFF$ be a Banach function space
  compactly and densely embedded in $C(\mathbf{X})$.
  Theorem \ref{thm:Banach}'s strategy guarantees,
  for all $N=1,2,\ldots$ and $F\in C(\mathbf{X})$,
  \begin{multline}\label{eq:continuous}
    \sum_{n=1}^N
    \left(
      y_n-\mu_n
    \right)^2
    \le
    \sum_{n=1}^N
    \left(
      y_n-F(x_n)
    \right)^2\\
    +
    C
    \inf_{\epsilon\in\left(0,1/2\right]}
    \left(
      \HHH_{\epsilon/A(\epsilon)}(U_{\FFF})
      +
      \log\log\frac{1}{\epsilon}
      +
      \log\log A(\epsilon)
      +
      \epsilon N
      +
      1
    \right),
  \end{multline}
  where $C$ is a universal constant and
  $
    A(\epsilon)
    :=
    2\max
    \bigl(
      1,
      \AAA_{\epsilon}^{\FFF}(F)
    \bigr)
  $.
\end{theorem}
\begin{proof}
  Inequality (\ref{eq:Banach}) immediately implies
  \begin{multline*}
    \sum_{n=1}^N
    \left(
      y_n-\mu_n
    \right)^2
    \le
    \sum_{n=1}^N
    \left(
      y_n-F(x_n)
    \right)^2\\
    +
    C
    \inf_{\delta>0}
    \inf_{\epsilon\in\left(0,1/2\right]}
    \left(
      \HHH_{\epsilon/A(\delta)}(U_{\FFF})
      +
      \log\log\frac{1}{\epsilon}
      +
      \log\log A(\delta)
      +
      \epsilon N
      +
      4\delta N
      +
      1
    \right),
  \end{multline*}
  and it remains to restrict $\delta$ to $\delta\in\left(0,1/2\right]$
  and set $\epsilon:=\delta$.
  \qedtext
\end{proof}

Theorem \ref{thm:continuous} will be the source of many universal prediction strategies.
Given any of the Banach spaces compactly and densely embedded in $C(\mathbf{X})$
introduced in \S\S\ref{sec:analytic}--\ref{sec:Sobolev},
Theorem \ref{thm:continuous} produces a universal prediction strategy:
it is clear that (\ref{eq:continuous}) implies (\ref{eq:asymptotic}).

\section{Finite-dimensional benchmark classes}
\label{sec:finite-dimensional}

We will be using (following \cite{\KolmogorovTikhomirov})
the notation $f\sim g$
to mean $\lim_{\epsilon\to0}(f(\epsilon)/g(\epsilon))=1$
and the notation $f\asymp g$
to mean $f=O(g)$ and $g=O(f)$ as $\epsilon\to0$,
where $f$ and $g$ are positive functions of $\epsilon>0$.

If the benchmark class $\FFF$ is finite-dimensional,
the typical rate of growth of its metric entropy is
\begin{equation}\label{eq:L1}
  \HHH_{\epsilon}(\FFF)
  \sim
  L\log\frac{1}{\epsilon},
\end{equation}
where $L$ is the ``metric dimension'' of $\FFF$.
This motivates the following corollaries of Theorems \ref{thm:compact} and \ref{thm:Banach},
respectively.

\begin{corollary}\label{cor:compact-finite}
  Suppose $\FFF$ is a compact set in $C(\mathbf{X})$
  such that
  \begin{equation}\label{eq:L2}
    L
    :=
    \limsup_{\epsilon\to0}
    \frac{\HHH_{\epsilon}(\FFF)}{\log\frac{1}{\epsilon}}
    \in
    (0,\infty).
  \end{equation}
  There exists a strategy for Predictor
  that guarantees, for all $F\in\FFF$,
  \begin{equation}\label{eq:compact-finite}
    \sum_{n=1}^N
    \left(
      y_n-\mu_n
    \right)^2
    \le
    \sum_{n=1}^N
    \left(
      y_n-F(x_n)
    \right)^2
    +
    C L\log N
  \end{equation}
  from some $N$ on,
  where $C$ is a universal constant.
\end{corollary}
\begin{proof}
  It suffices to set $\epsilon:=1/N$ in (\ref{eq:compact}).
  (And it is easy to check that this value of $\epsilon$
  extracts from (\ref{eq:compact}) an optimal, to within a constant factor, regret term.)
  \qedtext
\end{proof}

\begin{corollary}\label{cor:Banach-finite}
  Let $\FFF$ be a Banach space
  embedded in $C(\mathbf{X})$
  and
  \begin{equation}\label{eq:L-Banach-finite}
    L
    :=
    \limsup_{\epsilon\to0}
    \frac{\HHH_{\epsilon}(U_{\FFF})}{\log\frac{1}{\epsilon}}
    \in
    (0,\infty).
  \end{equation}
  There exists a strategy for Predictor
  that guarantees, for all $F\in\FFF$,
  \begin{equation}\label{eq:Banach-finite}
    \sum_{n=1}^N
    \left(
      y_n-\mu_n
    \right)^2
    \le
    \sum_{n=1}^N
    \left(
      y_n-F(x_n)
    \right)^2
    +
    CL
    \log N
  \end{equation}
  from some $N$ on,
  where $C$ is a universal constant.
\end{corollary}
Remember that any Banach spaces $\FFF$ satisfying (\ref{eq:L-Banach-finite})
is automatically finite-dimensional
(\cite{\KolmogorovTikhomirov}, Theorem XII).
\begin{Proof}{of Corollary \ref{cor:Banach-finite}}
  Since the Banach space $\FFF$ is finite-dimensional,
  it is compactly embedded in $C(\mathbf{X})$.
  Substituting $\epsilon:=1/N$ in (\ref{eq:Banach}),
  we obtain
  \begin{multline*}
    \sum_{n=1}^N
    \left(
      y_n-\mu_n
    \right)^2\\
    \le
    \sum_{n=1}^N
    \left(
      y_n-F(x_n)
    \right)^2
    +
    C
    \left(
      L\log(\phi N)
      +
      \log\log N
      +
      \log\log\phi
      +
      2
    \right)\\
    \le
    \sum_{n=1}^N
    \left(
      y_n-F(x_n)
    \right)^2
    +
    2CL\log N
  \end{multline*}
  from some $N$ on.
  \qedtext
\end{Proof}

Theorem \ref{thm:continuous} is irrelevant to this section:
no finite-dimensional subspace can be dense in $C(\mathbf{X})$
(since finite-dimensional subspaces are always closed).

\subsection*{Comparison with known results}

It is instructive to compare the bound of Corollary \ref{cor:Banach-finite}
with a standard bound in competitive linear regression,
obtained in \cite{vovk:2001competitive} for the prediction strategy
referred to as AAR in \cite{vovk:2001competitive}
and as the ``Vovk--Azoury--Warmuth forecaster'' in \cite{cesabianchi/lugosi:2006}.
In the metric entropy method the elements of a net in $\FFF$
(the union of $\epsilon$-nets of different balls in $\FFF$ for different $\epsilon$,
in the case of Theorem \ref{thm:Banach} and its corollaries) are AA mixed.
AAR is conceptually very similar:
instead of AA mixing the elements of the net,
it AA mixes $\FFF$ itself;
the weights assigned to the elements of the net
are replaced by a ``prior'' probability measure on $\FFF$,
and so summation is replaced by integration.
An advantage of this ``integration method''
is that, for a suitable choice of the prior measure,
it may produce a computationally efficient prediction strategy:
e.g., AAR, which uses a Gaussian measure as prior,
turned out to be a simple modification of ridge regression,
as computationally efficient as ridge regression itself.

Suppose that $\mathbf{X}$ is a bounded subset of $\bbbr^m$ and set
\begin{equation}\label{eq:Xs}
  X_2
  :=
  \sup_{x\in\mathbf{X}}
  \left\|
    x
  \right\|_2,
  \quad
  X_{\infty}
  :=
  \sup_{x\in\mathbf{X}}
  \left\|
    x
  \right\|_{\infty};
\end{equation}
it is clear that $X_2\le X_{\infty}$.
AAR guarantees
\begin{equation}\label{eq:AAR}
  \sum_{n=1}^N
  \left(
    y_n-\mu_n
  \right)^2
  \le
  \sum_{n=1}^N
  \left(
    y_n-\langle \theta,x_n\rangle
  \right)^2
  +
  \left\|
    \theta
  \right\|_2^2
  +
  m
  \ln
  \left(
    NX_{\infty}^2
    +
    1
  \right)
\end{equation}
(see \cite{vovk:2001competitive}, (22) with $a:=1$ and $Y^2$ replaced by $1$).
To extract a similar inequality from (\ref{eq:Banach-finite}),
let $U_m$ be the unit ball in $\bbbr^m$ equipped with the $\|\cdot\|_2$ norm,
$\FFF$ be the set of linear functions
$x\in\mathbf{X}\mapsto\langle\theta,x\rangle$, $\theta\in\bbbr^m$,
with the norm $\left\|\theta\right\|_2$,
and notice that
\begin{equation}\label{eq:entropy-finite}
  \HHH_{\epsilon}(U_{\FFF})
  \le
  X_2
  \HHH_{\epsilon}(U_{m})
  \le
  X_2
  \log
  \left\lceil
    \left(
      \frac{4}{\epsilon}
    \right)^m
  \right\rceil.
\end{equation}
The first inequality in (\ref{eq:entropy-finite})
follows from $X_2$ being the embedding constant of $\FFF$ into $C(\mathbf{X})$
(and also from the Cauchy--Schwarz inequality).
The second inequality in (\ref{eq:entropy-finite})
follows from the inequality (1.1.10) in \cite{carl/stephani:1990}.

\begin{remark*}
  A popular alternative
  (used in \cite{carl/stephani:1990}
  and, in a slightly modified form, \cite{edmunds/triebel:1996})
  to the notion of metric entropy
  $\HHH_{\epsilon}(A)$
  is that of \emph{entropy numbers} $\epsilon_n(A)$, $n=1,2,\ldots$,
  defined as the infimum of $\epsilon$
  such that there exists an $\epsilon$-net for $A$.
  Notice that the ``infimum'' here is attained
  (and so can be replaced by ``minimum'')
  because of the compactness of $A^n$.
  It is easy to see that
  \begin{equation}\label{eq:connection}
    2^{\HHH_{\epsilon}(A)}
    =
    \min
    \left\{
      n
      \st
      \epsilon_n(A)\le\epsilon
    \right\};
  \end{equation}
  this can be useful when translating results about entropy numbers
  into results about metric entropy.
\end{remark*}

Combining (\ref{eq:entropy-finite}) with Corollary \ref{cor:Banach-finite},
we obtain the following analogue of (\ref{eq:AAR}).

\begin{corollary}\label{cor:Banach-finite-VAW}
  Let $\mathbf{X}$ be a bounded set in $\bbbr^m$ and $X_2$ be defined by (\ref{eq:Xs}).
  There exists a strategy for Predictor
  that guarantees, for all $\theta\in\bbbr^m$,
  \begin{equation}\label{eq:Banach-finite-VAW}
    \sum_{n=1}^N
    \left(
      y_n-\mu_n
    \right)^2
    \le
    \sum_{n=1}^N
    \left(
      y_n-\langle \theta,x_n\rangle
    \right)^2
    +
    CX_2m
    \log N
  \end{equation}
  from some $N$ on,
  where $C$ is a universal constant.
\end{corollary}

An interesting feature of the regret terms in (\ref{eq:AAR})
and (\ref{eq:Banach-finite-VAW}) is their logarithmic dependence on $N$;
some other standard bounds,
such as those in \cite{cesabianchi/long/warmuth:1996},
\cite{kivinen/warmuth:1997} and \cite{auer/etal:2002},
involve $\sqrt{N}$
(or similar terms, such as the square root of the competitor's loss).
It is remarkable that the bound established in the first paper on competitive on-line regression,
\cite{foster:1991}, also depends on $N$ logarithmically;
the method used in that paper is penalized minimum least squares.
An important advantage of the bounds given in
\cite{cesabianchi/long/warmuth:1996,kivinen/warmuth:1997,auer/etal:2002}
is that the character of their dependence on the dimension $m$
allows one to carry them over to infinite-dimensional function spaces;
these bounds will be discussed again in \S\ref{sec:Sobolev}.

\section{Benchmark classes of analytic functions}
\label{sec:analytic}

In this section we consider classes of analytic functions,
and so it is natural to consider complex-valued functions
of one or more complex variables.
The observations are now any complex numbers,
$y_n\in\bbbc$,
bounded by $1$ in absolute value,
and so prediction rules are functions $F:\mathbf{X}\to\bbbc$.
Also,
in this section $C(\mathbf{X})$ will stand for the function space
of continuous complex-valued functions on $\mathbf{X}$.
It is clear that Theorems \ref{thm:asymptotic}--\ref{thm:continuous}
continue to hold in this extended framework.

According to \cite{\KolmogorovTikhomirov}, \S3.II,
the typical growth rate for the metric entropy
of infinite-dimensional classes $\FFF$
of analytic functions on $\mathbf{X}$ is
\begin{equation}\label{growth-analytic}
  \HHH_{\epsilon}(\FFF)
  \asymp
  \log^{m+1}\frac{1}{\epsilon},
\end{equation}
where $m$ is the dimension of $\mathbf{X}$.
(Although intermediate rates such as
\begin{equation*}
  \HHH_{\epsilon}(\FFF)
  \asymp
  \frac
  {
    \log^{m+1}\frac{1}{\epsilon}
  }
  {
    \log\log\frac{1}{\epsilon}
  }
\end{equation*}
also sometimes occur.)
For such growth rates (the complex versions of)
Theorems \ref{thm:compact}--\ref{thm:continuous}
imply the following three corollaries.

\begin{corollary}\label{cor:compact-analytic}
  Suppose $\FFF$ is a compact set in $C(\mathbf{X})$ and $M>0$ is such that
  \begin{equation*}
    L
    :=
    \limsup_{\epsilon\to0}
    \frac{\HHH_{\epsilon}(\FFF)}{\log^{M}\frac{1}{\epsilon}}
    \in
    (0,\infty).
  \end{equation*}
  There exists a strategy for Predictor
  that guarantees, for all $F\in\FFF$,
  \begin{equation}\label{eq:compact-analytic}
    \sum_{n=1}^N
    \left|
      y_n-\mu_n
    \right|^2
    \le
    \sum_{n=1}^N
    \left|
      y_n-F(x_n)
    \right|^2
    +
    CL\log^{M}N
  \end{equation}
  from some $N$ on,
  where $C$ is a universal constant.
\end{corollary}
\begin{proof}
  As in the proof of Corollary \ref{cor:compact-finite},
  set $\epsilon:=1/N$ in (\ref{eq:compact}).
  \qedtext
\end{proof}

\begin{corollary}\label{cor:Banach-analytic}
  Let $\FFF$ be a Banach function space
  compactly embedded in $C(\mathbf{X})$
  and $M>0$ be a number such that
  \begin{equation}\label{eq:L-and-M}
    L
    :=
    \limsup_{\epsilon\to0}
    \frac{\HHH_{\epsilon}(U_{\FFF})}{\log^{M}\frac{1}{\epsilon}}
    \in
    (0,\infty).
  \end{equation}
  There exists a strategy for Predictor
  that guarantees, for all $F\in\FFF$,
  \begin{equation}\label{eq:Banach-analytic}
    \sum_{n=1}^N
    \left|
      y_n-\mu_n
    \right|^2
    \le
    \sum_{n=1}^N
    \left|
      y_n-F(x_n)
    \right|^2
    +
    C L
    \log^{M}N
  \end{equation}
  from some $N$ on,
  where $C$ is a universal constant.
\end{corollary}
\begin{proof}
  Following the proof of Corollary \ref{cor:Banach-finite},
  we substitute $\epsilon:=1/N$ in (\ref{eq:Banach})
  to obtain, from some $N$ on:
  \begin{multline*}
    \sum_{n=1}^N
    \left|
      y_n-\mu_n
    \right|^2\\
    \le
    \sum_{n=1}^N
    \left|
      y_n-F(x_n)
    \right|^2
    +
    C
    \left(
      L\log^{M}(\phi N)
      +
      \log\log N
      +
      \log\log\phi
      +
      2
    \right)\\
    \le
    \sum_{n=1}^N
    \left|
      y_n-F(x_n)
    \right|^2
    +
    C' L
    \log^{M}N,
  \end{multline*}
  where $C'$ is another universal constant.
  \qedtext
\end{proof}

Unlike in the previous section,
Theorem \ref{thm:continuous} is not vacuous for classes of analytic functions:
as we will see in the following subsection,
there are numerous examples of such classes
that are compactly and densely embedded in $C(\mathbf{X})$,
for important signal spaces $\mathbf{X}$.
The following is the implication of Theorem \ref{thm:continuous}
for the growth rate (\ref{growth-analytic});
unfortunately,
this statement still has $\inf_{\epsilon}$
since the growth rate of $\AAA_{\epsilon}^{\FFF}(F)$ is unknown.

\begin{corollary}\label{cor:continuous-analytic}
  Let $\FFF$ be a Banach function space
  compactly and densely embedded in $C(\mathbf{X})$
  and let $L$ and $M$ be positive numbers satisfying (\ref{eq:L-and-M}).
  There exists a strategy for Predictor
  that guarantees, for all $F\in C(\mathbf{X})$,
  \begin{multline}\label{eq:continuous-analytic}
    \sum_{n=1}^N
    \left|
      y_n-\mu_n
    \right|^2
    \le
    \sum_{n=1}^N
    \left|
      y_n-F(x_n)
    \right|^2\\
    +
    C_M
    \inf_{\epsilon\in(0,1]}
    \left(
      L
      \left(
        \log^{+}
        \AAA_{\epsilon}^{\FFF}(F)
      \right)^M
      +
      L\log^M\frac{1}{\epsilon}
      +
      \epsilon N
    \right)
  \end{multline}
  from some $N$ on,
  where $C_M$ is a constant depending only on $M$
  and $\log^{+}$ is defined as
  \begin{equation*}
    \log^{+} t
    :=
    \begin{cases}
      \log t & \text{if $t\ge1$}\\
      0 & \text{otherwise}.
    \end{cases}
  \end{equation*}
\end{corollary}

\begin{proof}
  It is clear that the optimal value of $\epsilon$
  in the regret term in (\ref{eq:continuous})
  tends to $0$ as $N\to\infty$,
  and so the regret term can be bounded above by
  \begin{multline*}
    C
    \inf_{\epsilon\in\left(0,1/2\right]}
    \left(
      L
      \log^M
      \left(
        \frac{A(\epsilon)}{\epsilon}
      \right)
      +
      \log\log\frac{1}{\epsilon}
      +
      \log\log A(\epsilon)
      +
      \epsilon N
      +
      1
    \right)\\
    \le
    C'
    \inf_{\epsilon\in\left(0,1\right]}
    \left(
      L
      \left(
        \log^{+}\AAA_{\epsilon}^{\FFF}(F)
        +
        \log\frac{1}{\epsilon}
      \right)^M
      +
      \epsilon N
    \right)
  \end{multline*}
  from some $N$ on.
  (The case $F\in\FFF$ has to be considered separately.)
  \qedtext
\end{proof}

\subsection*{Examples}

We will reproduce two simple examples
from \cite{\KolmogorovTikhomirov};
for simplicity we only consider analytic functions of one complex variable
(although already \cite{\KolmogorovTikhomirov}
contains results making extension to several variables straightforward).
Remember that the set of all complex numbers is denoted $\bbbc$.

Let $K$ be a simply connected continuum in $\bbbc$
containing more than one point
and $G$ be a region (connected open set)
such that $K\subseteq G\subseteq\bbbc$.
The set of all complex-valued functions on $K$
that admit a bounded analytic continuation to $G$
is denoted $A^K_G$.
Equipped with the usual pointwise addition and scalar action
and with the norm
\begin{equation}\label{eq:norm-1}
  \left\|
    f|_K
  \right\|_{A^K_G}
  :=
  \sup_{z\in G}
  \left|
    f(z)
  \right|,
\end{equation}
where $f:G\to\bbbc$ ranges over the bounded analytic functions
and $f|_K$ is the restriction of $f$ to $K$,
it becomes a Banach space.
Expression (\ref{eq:norm-1}) is well-defined
by the uniqueness theorem in complex analysis,
and the completeness of $A^K_G$ follows from the fact
(known as Weierstrass's theorem, \cite{ahlfors:1979}, Theorem IV.1.1)
that uniform limits of analytic functions are analytic.

It is shown in \cite{\KolmogorovTikhomirov}, (139), that
\begin{equation}\label{eq:Kolmogorov}
  \HHH_{\epsilon}
  \left(
    U_{A_G^K}
  \right)
  \sim
  \tau(G,K)
  \log^2\frac{1}{\epsilon}
\end{equation}
(this was hypothesised by Kolmogorov and proved independently
by Babenko and Erokhin;
in \cite{widom:1972}
the constant $\tau(G,K)$ was shown,
under mild restrictions,
to be proportional to the Green capacity of $K$ relative to $G$).
Therefore,
Corollary \ref{cor:Banach-analytic}
gives a strategy for Predictor guaranteeing
\begin{equation}\label{eq:Banach-analytic-1}
  \sum_{n=1}^N
  \left|
    y_n-\mu_n
  \right|^2
  \le
  \sum_{n=1}^N
  \left|
    y_n-F(x_n)
  \right|^2
  +
  C\tau(G,K)
  \log^2 N
\end{equation}
for all $F\in A^K_G$ and from some $N$ on,
where $C$ is a universal constant.

In many interesting special cases considered in \cite{\KolmogorovTikhomirov},
\S7,
the constant $\tau(G,K)$ has a simple explicit expression, e.g.:
\begin{itemize}
\item
  $\tau(G,K)=1/\log(R/r)$ if $K=r\overline{\bbbd}$ and $G=R\bbbd$,
  $R>r>0$,
  $\bbbd:=U_{\bbbc}$ being the open unit disk in $\bbbc$;
\item
  $\tau(G,K)=1/(2\log\lambda)$ if $K=[-1,1]$
  and $G$ is the ellipse with the sum of semi-axes equal to $\lambda>1$
  and with foci at the points $\pm1$
  (there is a misprint in \cite{\KolmogorovTikhomirov}, (131);
  the correct formula is given in, e.g., \cite{\Vitushkin}, Theorem 1 in \S12).
\end{itemize}
Both these expressions were obtained by Vitushkin.

Let $h>0$.
The vector space of all periodic period $2\pi$ complex-valued functions on the real line $\bbbr$
that admit a bounded analytic continuation to the strip
$\left\{z\in\bbbc\st\left|\Im z\right|<h\right\}$
is denoted $A_h$.
The norm in this space is defined by
\begin{equation}\label{eq:norm-2}
  \left\|
    f|_{\bbbr}
  \right\|_{A_h}
  :=
  \sup_{z:\left|\Im z\right|<h}
  \left|
    f(z)
  \right|,
\end{equation}
where $f$ ranges over the bounded analytic functions
on $\left\{z\st\left|\Im z\right|<h\right\}$.
Expression (\ref{eq:norm-2}) is again well-defined
and the normed space $A_h$ is complete.
The estimate of the metric entropy of the unit ball of $A_h$
given in \cite{\KolmogorovTikhomirov}, (130), is
\begin{equation}\label{eq:Vitushkin}
  \HHH_{\epsilon}
  \left(
    U_{A_h}
  \right)
  \sim
  \frac{2}{h\log e}
  \log^2\frac{1}{\epsilon}
\end{equation}
(Vitushkin).
Corollary \ref{cor:Banach-analytic}
now gives
\begin{equation}\label{eq:Banach-analytic-2}
  \sum_{n=1}^N
  \left|
    y_n-\mu_n
  \right|^2
  \le
  \sum_{n=1}^N
  \left|
    y_n-F(x_n)
  \right|^2
  +
  \frac{C}{h}
  \log^2 N
\end{equation}
for all $F\in A_h$ and from some $N$ on,
where $C$ is a universal constant.

Corollary \ref{cor:continuous-analytic} is applicable to $\FFF=A_h$
for any $h>0$,
and so $A_h$ gives rise to a universal prediction strategy.
Indeed, taking $\mathbf{X}:=\partial\bbbd$
(the unit circle in $\bbbc$)
and identifying complex-valued functions on $\partial\bbbd$
with the corresponding periodic period $2\pi$ complex-valued functions on $\bbbr$
(namely, $f:\partial\bbbd\to\bbbc$ is identified
with the function $t\in\bbbr\mapsto f\left(e^{it}\right)$),
we can arbitrarily closely in $C(\mathbf{X})$
approximate each $F\in C(\mathbf{X})$ by a trigonometric polynomial
(this is Weierstrass's second theorem,
\cite{achieser:1956}, \S21),
whose analytic continuation to
$\left\{z\st\left|\Im z\right|<h\right\}$
is bounded;
we can see that $A_h$ is dense in $C(\mathbf{X})$.

Suppose $\mathbf{X}\subseteq\bbbc$ is compact
(in particular, closed).
For $A_G^{\mathbf{X}}$ to be dense in $C(\mathbf{X})$,
$\mathbf{X}$ must be nowhere dense in $\bbbc$
(since limits in $C(\mathbf{X})$ of elements of $A_G^{\mathbf{X}}$
would be analytic in the interior points of $\mathbf{X}$).
If we additionally assume that $\mathbf{X}$ is simply connected,
Mergelyan's theorem
(\cite{rudin:1987}, Theorem 20.5%
)
will guarantee that every continuous complex-valued function on $\mathbf{X}$
can be arbitrarily closely in $C(\mathbf{X})$
approximated by a polynomial.
We can see that $A_G^{\mathbf{X}}$ is dense in $C(\mathbf{X})$
provided $\mathbf{X}$ is a nowhere dense simply connected compact.
The most interesting case is perhaps where $\mathbf{X}=[a,b]$
is a closed interval in $\bbbr$.

\subsection*{Dense function spaces popular in learning theory}

Benchmark classes such as $A^K_G$ and $A_h$
have never been used, to my knowledge,
in competitive on-line prediction.
Familiar rates of growth of the regret term are $O(\log N)$ or $N^{\alpha}$
(for $\alpha\in(0,1)$, usually $\alpha=1/2$);
intermediate rates obtainable for $A^K_G$ and $A_h$,
such as (\ref{eq:Banach-analytic-1}) and (\ref{eq:Banach-analytic-2}),
have not been known.

Several benchmark classes of this type, however,
have been implicitly considered since they are reproducing kernel Hilbert spaces
corresponding to popular reproducing kernels
(see \cite{vapnik:1998} and \cite{scholkopf/smola:2002}
for the use of reproducing kernels in learning theory
and \cite{aronszajn:1950} for the theory of reproducing kernel Hilbert spaces,
or RKHS for brevity).
One of such spaces is the Hardy space $H^2(\bbbd)$
restricted to the interval $(-1,1)$ of the real line
(see, e.g., \cite{paulsen:2006}).
Mergelyan's theorem
(or Weierstrass's first theorem,
\cite{achieser:1956}, \S20)
immediately implies
that for each $\epsilon>0$ the restriction of $H^2(\bbbd)$ to $[-1+\epsilon,1-\epsilon]$
is dense in $C([-1+\epsilon,1-\epsilon])$:
indeed, each polynomial belongs to $H^2(\bbbd)$.
(In the multi-dimensional case,
this fact was established by Steinwart \cite{steinwart:2001},
Example 2.)
It is easy to see that, when $\mathbf{X}=[-1+\epsilon,1-\epsilon]$,
(\ref{eq:Banach-analytic-1}) holds not only for $A_G^K:=A_{\bbbd}^{\mathbf{X}}$
but also for $A_G^K$ replaced by the restriction of $H^2(\bbbd)$
to $\mathbf{X}$
and for $\tau(G,K)$ replaced by a suitable constant depending only on $\epsilon$.

\begin{remark*}
  The reproducing kernel
  \begin{equation*}
    \KKK(z,w)
    :=
    \frac{1}{1-\overline{w}z}
  \end{equation*}
  of the Hardy space $H^2(\bbbd)$ is known as the Szeg\"o kernel.
  In some recent learning literature
  (such as \cite{steinwart:2001}, Example 2,
  \cite{scholkopf/smola:2002}, Example 4.24)
  the restriction of the multidimensional Szeg\"o kernel
  to the unit ball in a Euclidean space
  is referred to as ``Vovk's infinite-degree polynomial kernel''.
  The origin of this undeserved name is the SVM manual \cite{saunders/etal:1998manual};
  I liked to use the Szeg\"o kernel
  when explaining the idea of reproducing kernels to my students.
\end{remark*}

Other popular spaces of analytic functions on $\bbbr^m$
are the reproducing kernel Hilbert spaces
corresponding to the ``Gaussian RBF kernels'',
parameterized by $\sigma>0$.
They are described in \cite{steinwart/etal:2004}
(and also earlier in \cite{bargmann:1961} and, more explicitly, \cite{saitoh:1997}).
We have for them both the $O(\log^2N)$ rate of growth of the regret term
and the denseness in $C(K)$ for each compact $K\subseteq\bbbr^m$
(see \cite{steinwart:2001}, Example 1).
Interestingly, these RKHS do not look dense in $C(K)$:
it appears that they can only approximate functions at the scale
comparable with the parameter $\sigma$
(perhaps the cause of this illusion
is the small metric entropy of these function classes).

In general,
it appears that most common reproducing kernels
give rise to RKHS consisting of analytic functions.
Suppose that $\mathbf{X}$ is a bounded set in a Euclidean space $\bbbr^m$.
It is often the case that the reproducing kernel $\KKK(z,w)$,
$z,w\in\mathbf{X}$,
admits a continuation
to a neighbourhood $O^2\subseteq(\bbbc^m)^2$ of $\overline{\mathbf{X}}^2$
analytic in its first argument $z$
and remaining a reproducing kernel.
By the Tietze--Uryson theorem
(\cite{engelking:1989}, 2.1.8)
there is an intermediate neighbourhood $G$,
such that $\overline{\mathbf{X}}\subseteq G\subseteq\overline{G}\subseteq O$.
Let $\FFF$ be the RKHS on $G$ generated by the given reproducing kernel $\KKK$
thus extended to $G^2$.
It is clear that
\begin{equation}\label{eq:c-first-occurrence}
  \ccc_{\FFF}
  :=
  \sup_{z\in G}
  \sqrt{\KKK(z,z)}
\end{equation}
is finite.
The set of the evaluation functionals $\KKK_w(z):=\KKK(z,w)$, $w\in G$,
is dense in $\FFF$
(\cite{aronszajn:1950}, \S2(4);
for details, see \cite{aronszajn:1943}, Theorem 2),
convergence in $\FFF$ implies convergence in $C(G)$
(by $\ccc_{\FFF}<\infty$ and \cite{aronszajn:1950}, \S2(5)),
each $\KKK_w$ is analytic,
and uniform limits of analytic functions are analytic
(\cite{ahlfors:1979}, Theorem IV.1.1);
therefore,
$\FFF$ consists of analytic functions.
Since $\ccc_{\FFF}<\infty$,
we have $U_{\FFF}\subseteq\ccc_{\FFF}U_{A_G^{\mathbf{X}}}$,
and so $\FFF$ is compactly embedded in $C(\mathbf{X})$
and, as above, the regret term grows as a polynomial of $\log N$\label{p:poly-log}.

Steinwart \cite{steinwart:2001} gives four examples
of reproducing kernels on $\mathbf{X}^2$
that can be analytically continued to a neighbourhood of $\mathbf{X}^2$,
as in the previous paragraph,
and whose RKHS are dense in $C(\mathbf{X})$
(we described the first two of his examples above).

Sometimes formulas for reproducing kernels contain ``awkward'' building blocks
such as taking the fractional part
(\cite{wahba:1990}, (10.2.4)),
absolute value,
or $\min$
(\cite{\DFIII}, (8)),
and in this case analytic continuation to a neighbourhood is usually impossible.
Such reproducing kernels are often derived from the corresponding RKHS
that are much more massive than the classes of analytic functions
considered in this section;
such massive classes will be considered in the next section.

\section{Sobolev-type classes}
\label{sec:Sobolev}

We now return to our basic prediction protocol
in which the observations $y_n$ are real numbers
(bounded by 1 in absolute value);
$C(\mathbf{X})$ will again denote the continuous real-valued functions on $\mathbf{X}$.

Typical classes studied in the theory of functions of real variable
are much more massive than typical classes of analytic functions.
In the second part of this section
we will see examples showing
that the typical growth rate for the metric entropy
of compact classes $\FFF$ of real-valued functions
defined on nice subsets of a Euclidean space is
\begin{equation}\label{eq:typical-growth-Sobolev}
  \HHH_{\epsilon}(\FFF)
  \asymp
  \left(
    1/\epsilon
  \right)^{\gamma},
\end{equation}
where $\gamma>0$ is the ``degree of non-smoothness'' of $\FFF$.
The following two corollaries are asymptotic versions of Theorems \ref{thm:compact}--\ref{thm:Banach}
for this growth rate.

\begin{corollary}\label{cor:compact-Sobolev}
  Suppose a compact set $\FFF$ in $C(\mathbf{X})$ and a positive number $\gamma$ satisfy
  \begin{equation*}
    L
    :=
    \limsup_{\epsilon\to0}
    \frac{\HHH_{\epsilon}(\FFF)}{(1/\epsilon)^{\gamma}}
    \in
    (0,\infty).
  \end{equation*}
  There exists a strategy for Predictor
  that guarantees, for all $F\in\FFF$,
  \begin{equation}\label{eq:compact-Sobolev}
    \sum_{n=1}^N
    \left(
      y_n-\mu_n
    \right)^2
    \le
    \sum_{n=1}^N
    \left(
      y_n-F(x_n)
    \right)^2
    +
    C
    L^{\frac{1}{\gamma+1}}
    N^{\frac{\gamma}{\gamma+1}}
  \end{equation}
  from some $N$ on,
  where $C$ is a universal constant.
\end{corollary}
\begin{proof}
  Solving
  \begin{equation}\label{eq:minimize}
    L(1/\epsilon)^{\gamma} + \epsilon N \to \min,
  \end{equation}
  we obtain
  \begin{equation}\label{eq:epsilon}
    \epsilon
    =
    \left(
      \frac{L\gamma}{N}
    \right)^{\frac{1}{\gamma+1}}
    \to
    0
    \quad
    (N\to\infty)
  \end{equation}
  and 
  \begin{equation}\label{eq:base}
    L(1/\epsilon)^{\gamma} + \epsilon N
    =
    \left(
      \gamma^{\frac{1}{\gamma+1}}
      +
      \gamma^{-\frac{\gamma}{\gamma+1}}
    \right)
    L^{\frac{1}{\gamma+1}}
    N^{\frac{\gamma}{\gamma+1}};
  \end{equation}
  since the first factor on the right-hand side of the last expression
  always belongs to $(1,2]$,
  it can be ignored.
  \qedtext
\end{proof}

\begin{remark*}
  We will have to find minima such as (\ref{eq:minimize})
  on several occasions,
  and for the future reference I will give the general result of the calculation for
  \begin{equation}\label{eq:min}
    A \epsilon^{-a}
    +
    B \epsilon^{b}
    \to
    \min,
  \end{equation}
  where $A,B,a,b$ are positive numbers
  and $\epsilon$ ranges over $(0,\infty)$.
  The minimum is attained at
  \begin{equation}\label{eq:precise-epsilon}
    \epsilon
    =
    \left(
      \frac{Aa}{Bb}
    \right)^{\frac{1}{a+b}}
  \end{equation}
  and is equal to
  \begin{equation}\label{eq:precise-min}
    \left(
      (a/b)^{\frac{b}{a+b}}
      +
      (b/a)^{\frac{a}{a+b}}
    \right)
    A^{\frac{b}{a+b}}
    B^{\frac{a}{a+b}}.
  \end{equation}
  Instead of finding the precise minimum in (\ref{eq:min}),
  it will usually be more convenient to approximate it
  by equating the two addends in (\ref{eq:min}),
  which gives
  \begin{equation}\label{eq:approx-epsilon}
    \epsilon
    =
    \left(
      A/B
    \right)^{\frac{1}{a+b}}
  \end{equation}
  and so gives the upper bound
  \begin{equation}\label{eq:approx-min}
    2
    A^{\frac{b}{a+b}}
    B^{\frac{a}{a+b}}
  \end{equation}
  for (\ref{eq:precise-min}).
\end{remark*}

\begin{corollary}\label{cor:Banach-Sobolev}
  Let $\FFF$ be a Banach function space
  compactly embedded in $C(\mathbf{X})$
  and $\gamma$ be a positive number such that
  \begin{equation}\label{eq:L-and-gamma}
    L
    :=
    \limsup_{\epsilon\to0}
    \frac{\HHH_{\epsilon}(U_{\FFF})}{(1/\epsilon)^\gamma}
    \in
    (0,\infty).
  \end{equation}
  There exists a strategy for Predictor
  that guarantees, for all $F\in\FFF$,
  \begin{equation}\label{eq:Banach-Sobolev}
    \sum_{n=1}^N
    \left(
      y_n-\mu_n
    \right)^2
    \le
    \sum_{n=1}^N
    \left(
      y_n-F(x_n)
    \right)^2
    +
    C
    L^{\frac{1}{\gamma+1}}
    \phi^{\frac{\gamma}{\gamma+1}}
    N^{\frac{\gamma}{\gamma+1}}
  \end{equation}
  from some $N$ on,
  where $C$ is a universal constant
  and $\phi$ is defined by (\ref{eq:phi}).
\end{corollary}
\begin{proof}
  Substituting $L\phi^{\gamma}$ for $L$ on the right-hand side of (\ref{eq:base})
  and ignoring the first factor on the right-hand side,
  we obtain
  \begin{equation*}
    \left(L\phi^{\gamma}\right)^{\frac{1}{\gamma+1}}
    N^{\frac{\gamma}{\gamma+1}}
    =
    L^{\frac{1}{\gamma+1}}
    \phi^{\frac{\gamma}{\gamma+1}}
    N^{\frac{\gamma}{\gamma+1}}.
    \qedmath
  \end{equation*}
\end{proof}

\subsection*{Examples}

We will say that a function $F$ defined on a metric space with metric $\rho$
is \emph{H\"older continuous of order $\alpha\in(0,1]$ with coefficient $c>0$}
if, for all $x$ and $x'$ in the domain of $F$,
$
  \left|F(x)-F(x')\right|
  \le
  c
  \rho^{\alpha}(x,x')
$.
If $\alpha=1$,
we will also say that $F$ is \emph{Lipschitzian with coefficient $c$}.

Let $\mathbf{X}$ be an $m$-dimensional (axes-parallel) parallelepiped.
Define $\FFF$ to be the class of real-valued functions on $\mathbf{X}$
that are bounded in $C(\mathbf{X})$ by a given constant
and whose $k$th partial derivatives exist
and are all H\"older continuous of order $\alpha$
with a given coefficient.
It is shown in \cite{\KolmogorovTikhomirov}, Theorem XIII, that
\renewcommand{\theequation}{\ref{eq:typical-growth-Sobolev}}
\begin{equation}\label{p:second-entry}
  \HHH_{\epsilon}(\FFF)
  \asymp
  \left(
    1/\epsilon
  \right)^{\gamma},
\end{equation}
\renewcommand{\theequation}{\arabic{equation}}%
\addtocounter{equation}{-1}%
where $\gamma:=m/s=m/(k+\alpha)$ is the ``degree of non-smoothness''
($1/\gamma$ was called the ``degree of smoothness''
by G.~G.~Lorentz in his review of \cite{\KolmogorovTikhomirov}
in \emph{Mathematical Reviews})
and $s:=k+\alpha$ is the ``indicator of smoothness''
(\ifnotLATIN\begin{cyr}pokazatel\cyrsftsn\ gladkosti\end{cyr}, \fi
\cite{\KolmogorovTikhomirov}, \S3.III).
We can now deduce from (\ref{eq:compact-Sobolev}) that
\begin{equation}\label{eq:specific}
  \sum_{n=1}^N
  \left(
    y_n-\mu_n
  \right)^2
  \le
  \sum_{n=1}^N
  \left(
    y_n-F(x_n)
  \right)^2
  +
  C_{\FFF}
  N^{\frac{m}{m+s}}
\end{equation}
for all $F\in\FFF$ and from some $N$ on,
where $C_{\FFF}$ is a constant depending on $\FFF$ but nothing else.

For the class $\FFF$ of Lipschitzian functions with coefficient $c$
defined on an interval of the real line of length $l$
and bounded in absolute value by a given constant
Kolmogorov and Tikhomirov \cite{\KolmogorovTikhomirov}
(see their (10), which also remains true
when $\HHH_{\epsilon}(A)$ is replaced by $\HHH_{\epsilon}^A(A)$)
obtain the more accurate estimate $\HHH_{\epsilon}(\FFF)\sim cl/\epsilon$.
In this case (\ref{eq:specific}) can be replaced by
\begin{equation}\label{eq:Lipschitz}
  \sum_{n=1}^N
  \left(
    y_n-\mu_n
  \right)^2
  \le
  \sum_{n=1}^N
  \left(
    y_n-F(x_n)
  \right)^2
  +
  C
  \sqrt{cl}
  N^{1/2},
\end{equation}
where $C$ is a universal constant.

Results of this type have been greatly extended in recent years.
We will later state one such result about Besov spaces $B^s_{p,q}(\mathbf{X})$%
.
For the general definition of Besov spaces
see \cite{edmunds/triebel:1996}, \S\S2.2,2.5.
Besov spaces $B^s_{p,q}$ are Banach spaces (assuming $p,q\ge1$),
but we will consider them as topological vector spaces
(i.e., will regard Banach spaces with equivalent norms
as the same space).

\begin{remark*}
  A popular definition of Besov spaces is via ``real interpolation''
  (as in \cite{adams/fournier:2003}, Chapter 7).
  For example, according to this definition,
  $B^s_{p,\infty}(\mathbf{X})$,
  where $s\in(0,\infty)$ and $p\in[1,\infty)$,
  consists of the functions $F$ whose Gagliardo set (\ref{eq:Gagliardo})
  with $C(\mathbf{X})$ replaced by $L^p(\mathbf{X})$
  and $\FFF$ replaced by the Sobolev space $W^{m,p}(\mathbf{X})$
  (see \cite{adams/fournier:2003}, Chapter 3)
  for some integer $m>s$
  contains the curve
  \begin{equation*}
    \left\{
      (t_0,t_1)\in\bbbr^2
      \stbig
      t_0^{1-\theta}
      t_1^{\theta}
      =
      c
    \right\},
    \quad
    \theta:=s/m,
  \end{equation*}
  for some positive $c$;
  the infimum of $c$ with this property is the norm of $F$
  in $B^s_{p,\infty}(\mathbf{X})$.
\end{remark*}

We are only interested in Besov spaces
whose domain is the signal space $\mathbf{X}$.
In the rest of this section it will always be assumed that $\mathbf{X}$
is a subset of Euclidean space,
$\mathbf{X}\subseteq\bbbr^m$,
which is a \emph{minimally regular domain},
in the sense that it is bounded and coincides with the interior of its closure
(\cite{edmunds/triebel:1996}, Definition 2.5.1/2).

Every $B^s_{p,q}(\mathbf{X})$ with $s>m/p$
is compactly embedded in $C(\mathbf{X})$
(apply \cite{edmunds/triebel:1996}, (2.5.1/10),
to $s_1:=s$, $p_1:=p$, $q_1:=q$, $p_2:=q_2:=\infty$ and sufficiently small $s_2>0$
and remember that $\CCC^s(\mathbf{X}):=B^s_{\infty,\infty}(\mathbf{X})$
are H\"older--Zygmund spaces,
\cite{edmunds/triebel:1996}, 2.2.2(iv)).
We will be interested only in this case.
Edmunds and Triebel's general result
(Theorem 3.5 of \cite{edmunds/triebel:1996}
applied to $s_1:=s$, $p_1:=p$, $q_1:=q$, $s_2:=0$, $p_2:=\infty$ and $q_2:=1$,
in combination with (2.3.3/3))
then shows that
\begin{equation*}
  \HHH_{\epsilon}
  \left(
    U_{B^s_{p,q}(\mathbf{X})}
  \right)
  \asymp
  (1/\epsilon)^{m/s}
\end{equation*}
(use (\ref{eq:connection}) to move between entropy numbers and metric entropy).
We can see that (\ref{eq:specific}) still holds
for $\FFF$ a bounded set in a general Besov space $B^s_{p,q}(\mathbf{X})$;
moreover,
Corollary \ref{cor:Banach-Sobolev} shows that
\begin{equation}\label{eq:Banach-Sobolev-Besov}
  \sum_{n=1}^N
  \left(
    y_n-\mu_n
  \right)^2
  \le
  \sum_{n=1}^N
  \left(
    y_n-F(x_n)
  \right)^2
  +
  C_{\mathbf{X},s,p,q}
  \left(
    \left\|
      F
    \right\|_{B^s_{p,q}(\mathbf{X})}
    +
    1
  \right)^{\frac{m}{m+s}}
  N^{\frac{m}{m+s}}
\end{equation}
for all $F\in B^s_{p,q}(\mathbf{X})$ from some $N$ on,
where $C_{\mathbf{X},s,p,q}$ is a constant depending only on $\mathbf{X},s,p,q$.
Setting $p$ and $q$ to $\infty$,
we recover (\ref{eq:specific}).

To conclude this subsection,
let us go back to reproducing kernels.
Cucker and Smale (\cite{cucker/smale:2002}, Theorem D)
show that if $\FFF$ is an RKHS with a $C^{\infty}$ reproducing kernel
on $\mathbf{X}^2$ for a compact set $\mathbf{X}\subseteq\bbbr^m$,
\begin{equation*}
  \HHH_{\epsilon}
  \left(
    U_{\FFF}
  \right)
  =
  O
  \left(
    (1/\epsilon)^{2m/h}
  \right)
\end{equation*}
for an arbitrary $h>m$.
Corollary \ref{cor:Banach-Sobolev}
(together with its proof,
since the $L$ in (\ref{eq:L-and-gamma}) is $0$ for each $\gamma$
and so has to be replaced by an upper bound)
shows that, for an arbitrarily small $\delta>0$,
\begin{equation}\label{eq:Banach-Sobolev-RKHS}
  \sum_{n=1}^N
  \left(
    y_n-\mu_n
  \right)^2
  \le
  \sum_{n=1}^N
  \left(
    y_n-F(x_n)
  \right)^2
  +
  N^{\delta}
\end{equation}
for all $F\in\FFF$ from some $N$ on.
The regret term in (\ref{eq:Banach-Sobolev-RKHS}) is not as good
as the poly-log regret term for RKHS with analytic reproducing kernels
(see p.~\pageref{p:poly-log}),
but this is not surprising:
the class of analytic functions is known to be much narrower
than that of infinitely differentiable functions
(for a useful relation between these classes
see \cite{timan:1963}, 3.7.1).

\subsection*{Comparisons with defensive forecasting}

Many of the Besov spaces $B^{s}_{p,q}(\mathbf{X})$ are ``uniformly convex'',
and this makes it possible to apply to them a result
obtained in \cite{\DFVI} using the method of ``defensive forecasting''.

Let $V$ be a Banach space and
$
  \partial U_{V}
  :=
  \left\{
    v\in V
    \st
    \left\|
      v
    \right\|_{V}
    =
    1
  \right\}
$
be the unit sphere in $V$
(the boundary of the unit ball $U_V$).
A convenient measure of rotundity of the unit ball $U_V$
is Clarkson's \cite{clarkson:1936} modulus of convexity
\begin{equation}\label{eq:clarkson}
  \delta_{U}(\epsilon)
  :=
  \inf_{\substack{u,v\in \partial U_{V}\\\left\|u-v\right\|_{V}=\epsilon}}
  \left(
    1
    -
    \left\|
      \frac{u+v}{2}
    \right\|_{V}
  \right),
  \quad
  \epsilon\in(0,2]
\end{equation}
(we will be mostly interested in the small values of $\epsilon$).

If a Banach space $\FFF$ is continuously embedded in $C(\mathbf{X})$,
the embedding constant will be denoted $\ccc_{\FFF}$:
\begin{equation}\label{eq:finiteness}
  \ccc_{\FFF}
  :=
  \sup_{F\in U_{\FFF}}
  \left\|
    F
  \right\|_{C(\mathbf{X})}
  <
  \infty
\end{equation}
(we have already used this notation
in the special case of RKHS:
cf.\ (\ref{eq:c-first-occurrence})).

\begin{proposition}[\cite{\DFVI}, Theorem 1]\label{prop:DFVI}
  Let $\FFF$ be a Banach space continuously embedded in $C(\mathbf{X})$
  and such that
  \begin{equation}\label{eq:condition}
    \forall\epsilon\in(0,2]:
    \delta_{\FFF}(\epsilon)\ge(\epsilon/2)^p/p
  \end{equation}
  for some $p\in[2,\infty)$.
  There exists a strategy for Predictor producing $\mu_n$
  that are guaranteed to satisfy
  \begin{equation}\label{eq:performance}
    \sum_{n=1}^N
    \left(
      y_n-\mu_n
    \right)^2
    \le
    \sum_{n=1}^N
    \left(
      y_n-F(x_n)
    \right)^2
    +
    40
    \sqrt{\ccc_{\FFF}^2+1}
    \left(
      \left\|
        F
      \right\|_{\FFF}
      +
      1
    \right)
    N^{1-1/p}
  \end{equation}
  for all $N=1,2,\ldots$ and all $F\in\FFF$.
\end{proposition}
It is interesting that in Proposition \ref{prop:DFVI}
$\FFF$ is not required to be compactly embedded in $C(\mathbf{X})$.

It was shown by Clarkson (\cite{clarkson:1936}, \S3) that, for $p\in[2,\infty)$,
\begin{equation*}
  \delta_{L^p}(\epsilon)
  \ge
  1
  -
  \left(
    1
    -
    (\epsilon/2)^p
  \right)^{1/p}.
\end{equation*}
(And this bound was shown to be optimal in \cite{hanner:1956}.)
%
This result was extended to some other Besov spaces
in \cite{cobos/edmunds:1988}, Theorem 3:
the modulus of convexity of each Besov space $B^{s}_{p,q}(\bbbr^m)$,
$s\in\bbbr$, $p\in[2,\infty)$
and $q\in[p/(p-1),p]$,
also satisfies
\begin{equation}\label{eq:delta-Besov}
  \delta_{B^{s}_{p,q}(\bbbr^m)}(\epsilon)
  \ge
  1
  -
  \left(
    1
    -
    (\epsilon/2)^p
  \right)^{1/p}.
\end{equation}

Edmunds and Triebel \cite{edmunds/triebel:1996}, 2.5.1,
define the Besov space $B^{s}_{p,q}(\mathbf{X})$ on $\mathbf{X}\subseteq\bbbr^m$
as the set of all restrictions of the functions in $B^{s}_{p,q}(\bbbr^m)$ to $\mathbf{X}$
with the norm
\begin{equation}\label{eq:norm-domain}
  \left\|
    F
  \right\|_{B^{s}_{p,q}(\mathbf{X})}
  :=
  \inf_{F^*}
  \left\|
    F^*
  \right\|_{B^{s}_{p,q}(\bbbr^m)},
\end{equation}
where $F^*$ ranges over all extensions of $F$ to $\bbbr^m$.
To check that $B^{s}_{p,q}(\mathbf{X})$
is at least as convex as $B^{s}_{p,q}(\bbbr^m)$
for $p\ge2$ and $p/(p-1)\le q\le p$,
take any $F_1,F_2\in B^{s}_{p,q}(\mathbf{X})$ of norm $1$
and at a distance of $\epsilon$ from each other.
If the infima in the definition (\ref{eq:norm-domain}) of the norms of $F_1$ and $F_2$
are attained,
we can take the extensions $F^*_1$ and $F^*_2$ to $\bbbr^m$ of norm $1$
and notice that,
as $\left\|F^*_1-F^*_2\right\|_{B^s_{p,q}(\bbbr^m)}\ge\epsilon$
and the modulus of convexity is a non-decreasing function of $\epsilon$
(\cite{lindenstrauss/tzafriri:1979}, Lemma 1.e.8),
\begin{equation*}
  \left\|
    \frac{F_1+F_2}{2}
  \right\|_{B^s_{p,q}(\mathbf{X})}
  \le
  \left\|
    \frac{F^*_1+F^*_2}{2}
  \right\|_{B^s_{p,q}(\bbbr^m)}
  \le
  1-\delta_{B^s_{p,q}(\bbbr^m)}(\epsilon).
\end{equation*}
If the infima are not attained,
we can still use a similar argument
for $p\ge2$ and $p/(p-1)\le q\le p$
with $\delta_{B^s_{p,q}(\bbbr^m)}$ replaced
by its lower bound given by (\ref{eq:delta-Besov}).
This shows that (\ref{eq:delta-Besov}) extends to arbitrary domains:
\begin{equation}\label{eq:delta-Besov-domain}
  \delta_{B^{s}_{p,q}(\mathbf{X})}(\epsilon)
  \ge
  1
  -
  \left(
    1
    -
    (\epsilon/2)^p
  \right)^{1/p}
  \ge
  (\epsilon/2)^p / p.
\end{equation}

Let $p\in[2,\infty)$,
$q\in[p/(p-1),p]$
and $s\in(m/p,\infty)$.
By Proposition \ref{prop:DFVI} and (\ref{eq:delta-Besov-domain}),
there exist a constant $C_{\mathbf{X},s,p,q}>0$
and a strategy for Predictor producing $\mu_n$
that are guaranteed to satisfy
\begin{equation}\label{eq:performance-DFVI}
  \sum_{n=1}^N
  \left(
    y_n-\mu_n
  \right)^2
  \le
  \sum_{n=1}^N
  \left(
    y_n-F(x_n)
  \right)^2
  +
  C_{\mathbf{X},s,p,q}
  \left(
    \left\|
      F
    \right\|_{B^s_{p,q}}
    +
    1
  \right)
  N^{1-1/p}
\end{equation}
for all $N=1,2,\ldots$ and all $F\in B^{s}_{p,q}(\mathbf{X})$.
We can see that defensive forecasting works better than metric entropy
at the ``wild'' end of the scale $B^s_{p,q}(\mathbf{X})$
whereas metric entropy better copes with smooth functions
(at this time we only pay attention to the exponent of $N$,
which is more important,
from the asymptotic point of view as $N\to\infty$,
than the coefficient in front of $N^{\cdots}$):
\begin{itemize}
\item
  Suppose $s\in(m/p,m/2]$.
  The exponent $1-1/p$ of $N$ in (\ref{eq:performance-DFVI})
  can be taken arbitrarily close to $1-s/m$,
  and we can see that it is then better than the exponent of $N$
  in (\ref{eq:Banach-Sobolev-Besov}):
  \begin{equation*}
    1 - \frac{s}{m}
    <
    \frac{m}{m+s}.
  \end{equation*}
  For example, in the very important case $m=1,s\approx1/2$
  (typical trajectories of the Brownian motion are of this type)
  defensive forecasting gives approximately $N^{1/2}$
  whereas the method of metric entropy gives approximately $N^{2/3}$.
\item
  Suppose $s\in(m/2,m)$.
  The exponent of $N$ in (\ref{eq:performance-DFVI})
  can always be taken as $1/2$,
  and it is still better than the exponent of $N$
  in (\ref{eq:Banach-Sobolev-Besov}):
  \begin{equation*}
    \frac{1}{2}
    <
    \frac{m}{m+s}.
  \end{equation*}
\item
  Suppose $s\in[m,\infty)$.
  A weakness of the method of defensive forecasting
  (in its current state: see, e.g., \cite{\DFV} and \cite{\DFVI},
  in addition to (\ref{eq:performance-DFVI}))
  is that it cannot give regret terms better than $O(N^{1/2})$.
  Therefore,
  the method of metric entropy beats defensive forecasting
  for smooth Besov spaces $B^s_{p,q}(\mathbf{X})$, $s>m$.
\end{itemize}

For comparison with (\ref{eq:specific}),
define the norm
\begin{equation}\label{eq:norm-k-alpha}
  \left\|
    F
  \right\|_{s}
  :=
  \max
  \left(
    \sup_{x\in\mathbf{X}}
    \left|
      F(x)
    \right|,
    \max_{\left|\beta\right|=k}
    \sup_{x,x'\in\mathbf{X}:x\ne x'}
    \frac
    {\left|D^{\beta}F(x)-D^{\beta}F(x')\right|}
    {\left\|x-x'\right\|^{\alpha}}
  \right),
\end{equation}
where $\mathbf{X}$ is a parallelepiped in $\bbbr^m$,
$\beta=(\beta_1,\ldots,\beta_m)$ ranges over the multi-indices,
$\left\|\cdot\right\|$ is any standard norm in $\bbbr^m$,
$F:\mathbf{X}\to\bbbr$ is $k$ times continuously differentiable function,
$\alpha\in(0,1]$,
and $s:=k+\alpha$.
It is easy to check that the Banach space normed by (\ref{eq:norm-k-alpha})
is continuously embedded in $B^{s'}_{p,2}(\mathbf{X})$ for any $s'<s$:
indeed,
it is obvious that the space normed by (\ref{eq:norm-k-alpha})
is continuously embedded in the H\"older--Zygmund space
$\CCC^s(\mathbf{X}):=B^s_{\infty,\infty}(\mathbf{X})$
(\cite{edmunds/triebel:1996}, 2.2.2(iv), \cite{triebel:1992}, 1.2.2),
and the usual embedding theorem implies that the H\"older--Zygmund space
is continuously embedded in $B^{s'}_{p,2}(\mathbf{X})$
(\cite{edmunds/triebel:1996}, (2.5.1/10), with $p_1=q_1=\infty$).
Fixing an arbitrarily small $\delta>0$,
we deduce from (\ref{eq:performance-DFVI})
that for each $s\le m/2$ there exists a constant $C_{\mathbf{X},s,\delta}>0$
such that
\begin{equation}\label{eq:performance-Holder-1}
  \sum_{n=1}^N
  \left(
    y_n-\mu_n
  \right)^2
  \le
  \sum_{n=1}^N
  \left(
    y_n-F(x_n)
  \right)^2
  +
  C_{\mathbf{X},s,\delta}
  \left(
    \left\|
      F
    \right\|_{s}
    +
    1
  \right)
  N^{1-s/m+\delta}
\end{equation}
for all $N=1,2,\ldots$ and all $F$ with finite $\left\|F\right\|_s$.
For $s$ above $m/2$ we have to take $p=2$
and so obtain
\begin{equation}\label{eq:performance-Holder-2}
  \sum_{n=1}^N
  \left(
    y_n-\mu_n
  \right)^2
  \le
  \sum_{n=1}^N
  \left(
    y_n-F(x_n)
  \right)^2
  +
  C_{\mathbf{X},s}
  \left(
    \left\|
      F
    \right\|_{s}
    +
    1
  \right)
  N^{1/2}
\end{equation}
in place of (\ref{eq:performance-Holder-1});
(\ref{eq:performance-Holder-2}) starts losing to (\ref{eq:specific})
when $s$ exceeds $m$.

It is also interesting to compare (\ref{eq:performance-Holder-2}) for $m=s=1$
with (\ref{eq:Lipschitz}).
Even though $\left\|F\right\|_{C(\mathbf{X})}\le1$ is the only interesting case,
the bound in (\ref{eq:Lipschitz}) still appears better:
it scales as $\surd c$ in $c$,
whereas (\ref{eq:performance-Holder-2}) scales as $c$
when applied to $\{F\st\left\|F\right\|_s\le c\}$.
This impression is confirmed by a more careful analysis:
(\ref{eq:Banach-Sobolev}) implies
\begin{equation*}
  \sum_{n=1}^N
  \left(
    y_n-\mu_n
  \right)^2
  \le
  \sum_{n=1}^N
  \left(
    y_n-F(x_n)
  \right)^2
  +
  C
  \sqrt
  {
    l
    \left(
      \left\|
        F
      \right\|_{s}
      +
      1
    \right)
    N
  }
\end{equation*}
for all $F$ with finite $\left\|F\right\|_s$ from some $N$ on,
where $C$ is a universal constant.
Comparing this with (\ref{eq:performance-Holder-2}),
we can see another disadvantage of defensive forecasting:
the regret term scales as the norm of $F$
(rather than its square root).

\subsection*{Other methods}

In this subsection I will briefly list some other methods
that have been used in competitive on-line regression.
It appears that the benchmark classes used have always belonged
to types I or III in the Kolmogorov--Tikhomirov classification.
This does not mean, however, that the available prediction algorithms
can be clearly divided into two groups corresponding to types I and III:
quite often an ostensibly type I algorithm can be easily extended
to benchmark classes that are infinite-dimensional Hilbert spaces
(of type III) using the so-called ``kernel trick''
(\cite{vapnik:1998}, \cite{scholkopf/smola:2002}).
Sometimes the possibility of such an extension is only stated
(more or less precisely)
without the actual extension being carried out.
In this subsection I will also discuss results of this type
(which might involve some conditions of regularity
that have not been stated explicitly).

Perhaps the most popular method for type III benchmark classes
is Gradient Descent,
together with its version,
Exponentiated Gradient
(the pioneering paper is \cite{cesabianchi/long/warmuth:1996};
see also \cite{kivinen/warmuth:1997} and \cite{auer/etal:2002}).
It is very efficient computationally
and often gives right orders of magnitude for the regret term.
As an example,
Auer \emph{et al.}\ (\cite{auer/etal:2002}, Theorem 3.1)
obtain, for their prediction algorithm using Gradient Descent,
the performance guarantee
\begin{multline}\label{eq:Auer}
  \sum_{n=1}^N
  (y_n-\mu_n)^2
  \le
  \sum_{n=1}^N
  (y_n-F(x_n))^2\\
  +
  8
  \ccc_{\FFF}^2
  c^2
  +
  8
  \ccc_{\FFF}
  c
  \sqrt
  {
    \frac12
    \sum_{n=1}^N
    (y_n-F(x_n))^2
    +
    \ccc_{\FFF}^2
    c^2
  }
\end{multline}
for all $N$ and all $F\in c U_{\FFF}$,
where $\FFF$ is a Hilbert space continuously embedded in $C(\mathbf{X})$
and $c$ is a known upper bound on $\left\|F\right\|_{\FFF}$.
The regret term is bounded above by
\begin{equation*}
  8
  \ccc_{\FFF}^2
  c^2
  +
  8
  \ccc_{\FFF}
  c
  \sqrt
  {
    2N
    +
    \ccc_{\FFF}^2
    c^2
  },
\end{equation*}
and so its growth rate is $O(N^{1/2})$;
this is typical for all popular methods for type III benchmark classes.
For comparison,
(\ref{eq:performance}) holds with $40$ replaced by $2$ when $p=2$
(\cite{\DFV}, Theorem 1).

Bounds involving the loss of the competitor in place of $N$,
such as (\ref{eq:Auer}),
have a clear advantage in situations where some competitors perform very well.
Such bounds can also be obtained using defensive forecasting
(see \cite{\DFV}, Theorem 2).

AAR can also be carried over to Hilbert spaces
(with the crucial step made in \cite{gammerman/etal:2004UAI}).
It gives a performance bound similar to (\ref{eq:performance}) with $p=2$,
but $40\sqrt{\ccc_{\FFF}^2+1}$ replaced by $2\ccc_{\FFF}$ 
(\cite{\DFV}, Theorem 3).
A simple example adapted from \cite{cesabianchi/long/warmuth:1996}
shows that the leading constant $2\ccc_{\FFF}$
cannot be decreased further
(\cite{\DFV}, Theorem 4).
(In general,
attention to the constants
is a tradition in learning theory
that distinguishes it from some parts of the theory of function spaces;
probably the impetus is coming from experimental machine learning
with its common struggle for small improvements in the performance
of prediction algorithms.)

\section{Very big classes}
\label{sec:image}

In this short section we will see an example
of a very fast growth rate of the regret term,
barely below the useless rate of $N$.
This slow rate is achieved not because of the richness of the function class $\FFF$
(as in \S\ref{sec:Sobolev} as compared to \S\ref{sec:analytic})
but because of the richness of the signal space $\mathbf{X}$ itself.

The corollary of this section is rather specialized.

\begin{corollary}\label{cor:compact-image}
  Suppose $\mathbf{X}$ is a totally bounded metric space
  and $\gamma$ is a positive number that satisfy
  \begin{equation}\label{eq:L}
    \HHH_{\epsilon}(\mathbf{X})
    \asymp
    (1/\epsilon)^{\gamma},
    \quad
    \epsilon\to0.
  \end{equation}
  Let $\FFF\subseteq C(\mathbf{X})$ consist of the H\"older continuous functions
  of order $\beta\in(0,1]$ with coefficient $c>0$
  that are bounded in absolute value by a given constant.
  There exists a strategy for Predictor
  that guarantees, for all $F\in\FFF$,
  \begin{equation}\label{eq:compact-image}
    \sum_{n=1}^N
    \left(
      y_n-\mu_n
    \right)^2
    \le
    \sum_{n=1}^N
    \left(
      y_n-F(x_n)
    \right)^2
    +
    C_{\beta,\gamma}
    c
    N / \log^{\beta/\gamma} N
  \end{equation}
  from some $N$ on,
  where $C_{\beta,\gamma}$ is a constant
  depending only on $\beta$ and $\gamma$.
\end{corollary}
\begin{proof}
  The modulus of continuity of $F$ is $\omega(\epsilon)\le c\epsilon^{\beta}$,
  and so $\omega^{-1}(\epsilon)\ge(\epsilon/c)^{1/\beta}$.
  Substituting this in Theorem XXV (more precisely, (233)) of \cite{\KolmogorovTikhomirov},
  we have
  \begin{equation*}
    \log \HHH_{\epsilon}(\FFF)
    =
    O
    \left(
      \HHH_{(\epsilon/2c)^{1/\beta}/2}(\mathbf{X})
    \right),
  \end{equation*}
  which in combination with (\ref{eq:L}) gives, for small enough $\epsilon>0$,
  \begin{equation*}
    \HHH_{\epsilon}(\FFF)
    \le
    2^{C_{\beta,\gamma}(c/\epsilon)^{\gamma/\beta}}.
  \end{equation*}
  Let $f(\epsilon)$ be the right-hand side of the last inequality.
  To estimate the infimum in (\ref{eq:compact}),
  we find $\epsilon$ from
  \begin{equation*}
    2^{C_{\beta,\gamma}(c/\epsilon)^{\gamma/\beta}}
    =
    N^{1/2}
  \end{equation*}
  (taking $N$ instead of $N^{1/2}$ would not improve $\epsilon$
  by more than a constant factor),
  which gives
  \begin{equation*}
    \epsilon
    =
    c
    \left(
      \frac{C_{\beta,\gamma}}{\frac12 \log N}
    \right)^{\beta/\gamma}
  \end{equation*}
  and the upper bound
  \begin{equation*}
    C'_{\beta,\gamma}
    cN
    \left(
      \frac{1}{\log N}
    \right)^{\beta/\gamma}
  \end{equation*}
  for the infimum in (\ref{eq:compact}),
  where $C'_{\beta,\gamma}$ is another constant
  depending only on $\beta$ and $\gamma$.
  \qedtext
\end{proof}

In view of (\ref{eq:typical-growth-Sobolev}) on p.~\pageref{p:second-entry}
we can take $\mathbf{X}$ to be the class of real-valued functions
on a parallelepiped in a Euclidean space
that are bounded in absolute value by a given constant
and whose $k$th partial derivatives exist
and are all H\"older continuous of order $\alpha$
with a given coefficient.
The signal space $\mathbf{X}$ can now be interpreted as the set of images
(admittedly, not very good images, without sharp boundaries between different objects).

\section{The role of the norm}
\label{sec:norm}

Our Theorems \ref{thm:compact}--\ref{thm:continuous}
in \S\ref{sec:entropy}
cover all values of $N$,
but starting from \S\ref{sec:finite-dimensional}
we switched to stating inequalities that hold from some $N$ on.
This allowed us to simplify the statements
and to tune our bounds to various parameters
of the considered benchmark classes.
On the negative side, however,
some important information was lost:
for example,
the inequality (\ref{eq:Banach-analytic})
does not involve the norm $\left\|F\right\|_{\FFF}$ of $F$
(whereas (\ref{eq:Banach-Sobolev}) retains the information about the norm).
The reason is that asymptotically, as $N\to\infty$,
the effect of $\left\|F\right\|_{\FFF}$ becomes negligible.
This is only true, however, if we fix $F$ while letting $N\to\infty$,
and it can be argued that this is not the only interesting asymptotics.
For example,
in the experimental machine learning,
$N$ is often a constant
(the size of the given data set)
and it is the norm $\left\|F\right\|_{\FFF}$ of the contemplated prediction rule $F$
that varies.
Another example will be provided by the considerations of the next section,
where the norm will be chosen as a function of $N$.
In this section we will discuss what happens if all
(or all but one)
values of $N$ are taken into account.
Interestingly,
this will change significantly our comparative evaluation of virtues
of some methods.

\subsection*{Finite-dimensional benchmark classes}

Instead of Corollary \ref{cor:Banach-finite} we now have:

\renewcommand{\thecorollary}{\ref{cor:Banach-finite}$^*$}
\begin{corollary}
  Let $\FFF$ be a finite-dimensional Banach space
  embedded in $C(\mathbf{X})$
  and $L\ge1$ be a number such that
  \begin{equation*}
    \HHH_{\epsilon}(U_{\FFF})
    \le
    L\log\frac{1}{\epsilon}
  \end{equation*}
  for all $\epsilon\in\left(0,1/2\right]$.
  There exists a strategy for Predictor
  that guarantees, for all $N=2,3,\ldots$ and all $F\in\FFF$,
  \begin{equation}\label{eq:Banach-finite-norm}
    \sum_{n=1}^N
    \left(
      y_n-\mu_n
    \right)^2
    \le
    \sum_{n=1}^N
    \left(
      y_n-F(x_n)
    \right)^2
    +
    C L
    \left(
      \log^{+}
      \left\|
        F
      \right\|_{\FFF}
      +
      \log N
    \right),
  \end{equation}
  where $C$ is a universal constant.
\end{corollary}
\addtocounter{corollary}{-1}%

\begin{proof}
  Substituting $\epsilon:=1/N$ in (\ref{eq:Banach}),
  we obtain:
  \begin{multline*}
    \sum_{n=1}^N
    \left(
      y_n-\mu_n
    \right)^2\\
    \le
    \sum_{n=1}^N
    \left(
      y_n-F(x_n)
    \right)^2
    +
    C
    \left(
      L\log(\phi N)
      +
      \log\log N
      +
      \log\log\phi
      +
      2
    \right)\\
    \le
    \sum_{n=1}^N
    \left(
      y_n-F(x_n)
    \right)^2
    +
    C
    \left(
      2L\log\phi
      +
      2L\log N
      +
      2
    \right),
  \end{multline*}
  which gives (\ref{eq:Banach-finite-norm})
  (for a different $C$).
  \qedtext
\end{proof}

\noindent
Instead of Corollary \ref{cor:Banach-finite-VAW}:

\renewcommand{\thecorollary}{\ref{cor:Banach-finite-VAW}$^*$}
\begin{corollary}
  Suppose $\mathbf{X}$ is a bounded set in $\bbbr^m$ and $X_2m\ge1$.
  There exists a strategy for Predictor
  that guarantees, for all $N=2,3,\ldots$ and all $\theta\in\bbbr^m$,
  \begin{equation}\label{eq:Banach-finite-VAW-norm}
    \sum_{n=1}^N
    \left(
      y_n-\mu_n
    \right)^2
    \le
    \sum_{n=1}^N
    \left(
      y_n-\langle \theta,x_n\rangle
    \right)^2
    +
    CX_2m
    \left(
      \log^{+}
      \left\|
        \theta
      \right\|_2
      +
      \log N
    \right),
  \end{equation}
  where $C$ is a universal constant.
\end{corollary}
\addtocounter{corollary}{-1}%

The coefficients in front of $\log N$
in the bounds (\ref{eq:AAR}) and (\ref{eq:Banach-finite-VAW-norm})
are not so different,
$m$ vs.\ $X_2m$
(ignoring the multiplicative constants).
The dependence on $\left\|\theta\right\|_2$ is, however, very different:
$\left\|\theta\right\|_2^2$ vs.\ $X_2m\log^{+}\left\|\theta\right\|_2$,
quadratic in (\ref{eq:AAR}) and logarithmic in (\ref{eq:Banach-finite-VAW-norm}).
The explanation is that AAR uses a Gaussian prior,
and so the weights assigned to remote $\theta$ decay very fast,
whereas in the method of metric entropy we used slowly decaying weights.
The quadratic dependence on $\left\|\theta\right\|_2$ is the price
that AAR pays for computational efficiency
(the former can be improved if AAR for different values of $a$ are AA mixed,
as in \cite{\DFV}, \S8,
but the latter might suffer).

We can see that the relation between the AAR bound
and the bound obtained using metric entropy
is not as straightforward as it seemed in \S\ref{sec:finite-dimensional}.
In fact,
the bounds are incomparable:
among the advantages of (\ref{eq:AAR}) are its explicitness,
a better coefficient in front of $\log N$,
and the simplicity and efficiency of the underlying prediction strategy;
however, the dependence of (\ref{eq:Banach-finite-VAW-norm})
on the norm of the competitor $\theta$ is better.

\subsection*{Benchmark classes of analytic functions}

In this subsection we will be using the conventions of \S\ref{sec:analytic};
in particular,
$C(\mathbf{X})$ will be the class of continuous complex-valued functions on $\mathbf{X}$.
Instead of Corollary \ref{cor:Banach-analytic} we have:

\renewcommand{\thecorollary}{\ref{cor:Banach-analytic}$^*$}
\begin{corollary}
  Let $\FFF$ be a Banach function space
  compactly embedded in $C(\mathbf{X})$
  and $L,M\in[1,\infty)$ be numbers such that
  \begin{equation}\label{eq:L-and-M-norm}
    \HHH_{\epsilon}(U_{\FFF})
    \le
    L
    \log^{M}\frac{1}{\epsilon}
  \end{equation}
  for all $\epsilon\in\left(0,1/2\right]$.
  There exists a strategy for Predictor
  that guarantees, for all $N=2,3,\ldots$ and all $F\in\FFF$,
  \begin{equation}\label{eq:Banach-analytic-norm}
    \sum_{n=1}^N
    \left|
      y_n-\mu_n
    \right|^2
    \le
    \sum_{n=1}^N
    \left|
      y_n-F(x_n)
    \right|^2
    +
    C_M L
    \left(
      \log^{+}
      \left\|
        F
      \right\|_{\FFF}
      +
      \log N
    \right)^{M},
  \end{equation}
  where $C_M$ is a constant depending only on $M$.
\end{corollary}
\addtocounter{corollary}{-1}%

\begin{proof}
  Substituting $\epsilon:=1/N$ in the complex version of (\ref{eq:Banach}),
  \begin{multline*}
    \sum_{n=1}^N
    \left|
      y_n-\mu_n
    \right|^2\\
    \le
    \sum_{n=1}^N
    \left|
      y_n-F(x_n)
    \right|^2
    +
    C
    \left(
      L\log^{M}(\phi N)
      +
      \log\log N
      +
      \log\log\phi
      +
      2
    \right)\\
    \le
    \sum_{n=1}^N
    \left|
      y_n-F(x_n)
    \right|^2
    +
    C'L
    \left(
      \log\phi
      +
      \log N
    \right)^{M},
  \end{multline*}
  where $C'$ is another universal constant.
  \qedtext
\end{proof}

Using Corollary \ref{cor:Banach-analytic}$^*$
instead of Corollary \ref{cor:Banach-analytic}
gives a strategy for Predictor guaranteeing,
instead of (\ref{eq:Banach-analytic-1}),
\begin{equation}\label{eq:Banach-analytic-1-norm}
  \sum_{n=1}^N
  \left|
    y_n-\mu_n
  \right|^2
  \le
  \sum_{n=1}^N
  \left|
    y_n-F(x_n)
  \right|^2
  +
  C_{G,K}
  \left(
    \log^{+}
    \left\|
      F
    \right\|_{A^K_G}
    +
    \log N
  \right)^2
\end{equation}
for all $F\in A^K_G$ and all $N=2,3,\ldots$,
where $C_{G,K}$ is a constant depending on $G$ and $K$ only.
Similarly, instead of (\ref{eq:Banach-analytic-2}) we have
\begin{equation}\label{eq:Banach-analytic-2-norm}
  \sum_{n=1}^N
  \left|
    y_n-\mu_n
  \right|^2
  \le
  \sum_{n=1}^N
  \left|
    y_n-F(x_n)
  \right|^2
  +
  C_h
  \left(
    \log^{+}
    \left\|
      F
    \right\|_{A_h}
    +
    \log N
  \right)^2
\end{equation}
for all $F\in A_h$ and all $N=2,3,\ldots$,
where $C_h$ is a constant depending on $h$ only.
Notice that the asymptotic expressions (\ref{eq:Kolmogorov}) and (\ref{eq:Vitushkin})
\emph{per se} do not provide any information on the dependence of $C_{G,K}$ on $G$ and $K$
and the dependence of $C_h$ on $h$.

\subsection*{Sobolev-type classes}

Instead of Corollary \ref{cor:Banach-Sobolev} we now have:

\renewcommand{\thecorollary}{\ref{cor:Banach-Sobolev}$^*$}
\begin{corollary}
  Let $\FFF$ be a Banach function space
  compactly embedded in $C(\mathbf{X})$
  and $L\ge 1$, $\gamma>0$ be numbers satisfying
  \begin{equation}\label{eq:L-and-gamma-norm}
    \HHH_{\epsilon}(U_{\FFF})
    \le
    L
    (1/\epsilon)^{\gamma}
  \end{equation}
  for all $\epsilon\in\left(0,1/2\right]$.
  There exists a strategy for Predictor
  that guarantees, for all $N=1,2,\ldots$ and all $F\in\FFF$,
  \begin{multline}\label{eq:Banach-Sobolev-norm}
    \sum_{n=1}^N
    \left(
      y_n-\mu_n
    \right)^2
    \le
    \sum_{n=1}^N
    \left(
      y_n-F(x_n)
    \right)^2\\
    +
    C
    \left(
      L^{\frac{1}{\gamma+1}}
      \phi^{\frac{\gamma}{\gamma+1}}
      N^{\frac{\gamma}{\gamma+1}}
      +
      \log^{+} \log \frac{N}{\gamma}
      +
      \log\log\phi
    \right),
  \end{multline}
  where $C$ is a universal constant
  and $\phi$ is defined by (\ref{eq:phi}).
\end{corollary}
\addtocounter{corollary}{-1}%

\begin{proof}
  See the proof of Corollary \ref{cor:Banach-Sobolev}
  (except that we cannot longer ignore the $\log\log$ terms in (\ref{eq:Banach})).
  The only case that remains to be considered is where $N$ is so small
  that $\epsilon$ in (\ref{eq:epsilon})
  (with $L$ replaced by $L\phi^{\gamma}$)
  fails to belong to $(0,1/2]$.
  In this case, however, the regret term of (\ref{eq:Banach-Sobolev-norm})
  exceeds $N/2$ because of the term $\epsilon N$ in (\ref{eq:Banach}),
  and so we can take $C:=2$.
  \qedtext
\end{proof}

We will refrain from stating
the non-asymptotic versions of the inequalities derived in \S\ref{sec:Sobolev}
for specific function classes:
such versions would be awkward
and would add little to our understanding
of the dependence of the regret term on the competitor's norm.

\section{Super-universal prediction?}
\label{sec:super}
\renewcommand{\thecorollary}{\arabic{corollary}}

In \S\S\ref{sec:analytic}--\ref{sec:Sobolev}
we dealt with universal prediction in the following,
somewhat vague (as most of our informal discussion in this section),
sense:
for a wide (in any case, dense in $C(\mathbf{X}$))
class $\FFF$ of continuous prediction rules
find a prediction strategy competitive with all $F\in\FFF$.
Possible dense classes $\FFF$ can be of very different size
even when defined on the same domain $\mathbf{X}$.
Even such meagre
(barely infinite-dimensional
from the point of view of metric entropy)
function classes as $A_G^K$ and $A_h$ of \S\ref{sec:analytic} are dense
(and so lead to a universal prediction strategy,
in the sense of Theorem \ref{thm:asymptotic}).
The classes of \S\ref{sec:Sobolev} are much larger.
However, we never know in advance which class $\FFF$ will work best
for our data sequence $x_1,y_1,x_2,y_2,\ldots$;
it would be ideal to have a prediction strategy
that works well for many different $\FFF$ simultaneously.
The study of existence of such ``super-universal'' prediction strategies
is a vast understudied (and ill-defined) area,
and in this section I will only make several simple and random observations.

\begin{remark*}
  There is a cheap way of achieving ``super-universality'':
  we can AA mix prediction strategies corresponding
  to many different classes $\FFF$.
  This would, however, further impair computational efficiency
  and possibly lead to cumbersome performance guarantees
  (remember that the classes we are interested in,
  such as $A_h$, $A_G^K$, $B^s_{p,q}$,
  often depend on one or more parameters).
\end{remark*}

Let us say that a function class $\FFF_1\subseteq C(\mathbf{X})$
``dominates'' a function class $\FFF_2\subseteq C(\mathbf{X})$
if any prediction strategy that performs not much worse
than the best small-norm prediction rules in $\FFF_1$
automatically performs not much worse
than the best small-norm prediction rules in $\FFF_2$.
(The corresponding definition for the case where $\FFF_1$ and $\FFF_2$
are compact classes of functions is simpler:
we can ignore the ``small-norm'' qualification.)
We will not try to formalize this ``definition'' in this paper.

Since we do not have any lower bounds in this paper,
when discussing the relation of domination
we will be comparing the available performance guarantees
rather than the optimal ones.
Hopefully, this will be corrected in the future work.

Ideally, there would be one or very few classes $\FFF$
that would dominate numerous other natural classes.
We will see in this section that less massive classes
often dominate more massive ones
(of course, with all the qualifications mentioned above).
There is no hope for finite-dimensional classes to dominate
infinite-dimensional ones,
and in the three subsections of this section
we will discuss the relation of domination between type II classes
and between type III classes,
and to what degree type II can dominate type III.

\subsection*{Domination between some classes of analytic functions}

In this and following subsections,
unlike \S\ref{sec:analytic},
we will consider periodic period $2\pi$ real-valued functions on $\bbbr$;
the function space $A_h$ is now defined as the class of all such functions
that can be analytically continued to $\{z\st\left|\Im z\right|\le h\}$,
with the norm defined to be the supremum norm of the analytic continuation
(which is unique).

We will be interested in the quality of competition with prediction rules
in $A_h$ achieved by the prediction strategy
designed for competing with prediction rules in $A_H$ for $H>h$.
But first we prove an auxiliary result.

\begin{lemma}\label{lem:analytic-by-analytic}
  Let $0<h<H<\infty$ and let $F\in A_h$.
  For small enough $\epsilon>0$,
  \begin{equation}\label{eq:analytic-by-analytic}
    \log\AAA_{\epsilon}^{A_H}(F)
    \le
    C
    \frac{H}{h}
    \log\frac{1}{\epsilon},
  \end{equation}
  where $C$ is a universal constant.
\end{lemma}

\begin{proof}
  According to Achieser's theorem
  (\cite{timan:1963}, 5.7.21; \cite{achieser:1956}, \S94)
  for sufficiently large $J$
  there is a trigonometric polynomial of degree $J$ at a uniform distance
  from $F$ at most
  \begin{equation*}
    \frac{8 c}{\pi}
    e^{-hJ}
  \end{equation*}
  where $c:=\left\|F\right\|_{A_h}$.
  To make sure that this does not exceed $\epsilon>0$
  (assumed sufficiently small),
  it suffices to set
  \begin{equation}\label{eq:J-1}
    J
    :=
    \left\lceil
      \frac{1}{h}
      \ln\frac{8c}{\pi\epsilon}
    \right\rceil.
  \end{equation}
  The absolute value of the approximating trigonometric polynomial
  does not exceed $\left\|F\right\|_{C(\bbbr)}+\epsilon$ on the real line
  and so does not exceed
  \begin{equation}\label{eq:max}
    \left(
      \left\|
        F
      \right\|_{C(\bbbr)}
      +
      \epsilon
    \right)
    e^{JH}
  \end{equation}
  in the strip $\left|\Im z\right|<H$
  (this follows from the Phragm\'en--Lindel\"of theorem:
  see \cite{timan:1963}, p.~13, footnote ${}^{****}$).
  Substituting (\ref{eq:J-1}) into (\ref{eq:max}),
  we find
  \begin{equation*}
    \log\AAA_{\epsilon}^{A_H}(F)
    \le
    \log
    \left(
      \left\|
        F
      \right\|_{C(\bbbr)}
      +
      \epsilon
    \right)
    +
    (\log e)
    JH
    \le
    C
    \frac{H}{h}
    \log\frac{1}{\epsilon}
  \end{equation*}
  for small enough $\epsilon>0$.
  \qedtext
\end{proof}

Combining Lemma \ref{lem:analytic-by-analytic}
with Theorem \ref{thm:continuous},
we obtain the following corollary.

\begin{corollary}\label{cor:analytic-from-analytic}
  Let $0<h<H<\infty$.
  The strategy for Predictor constructed in \S\ref{sec:analytic}
  for the benchmark class $A_H$ guarantees
  \begin{equation}\label{eq:analytic-from-analytic}
    \sum_{n=1}^N
    \left(
      y_n-\mu_n
    \right)^2
    \le
    \sum_{n=1}^N
    \left(
      y_n-F(x_n)
    \right)^2
    +
    C
    \frac{H^2}{h^3}
    \log^2 N
  \end{equation}
  for each $F\in A_h$ from some $N$ on,
  where $C$ is a universal constant.
\end{corollary}

\begin{proof}
  The regret term in (\ref{eq:continuous-analytic})
  can be bounded above by
  \begin{multline*}
    C'_M
    \left[
      L
      \left(
        \frac{H}{h}
        \log\frac{1}{\epsilon}
      \right)^{M}
      +
      L \log^2\frac{1}{\epsilon}
      +
      \epsilon N
    \right]_{\epsilon=1/N}\\
    \le
    C
    \left[
      \frac{H^2}{h^3}
      \log^2\frac{1}{\epsilon}
      +
      \epsilon N
    \right]_{\epsilon=1/N}
    \le
    C'
    \frac{H^2}{h^3}
    \log^2 N.
  \end{multline*}
  In this chain,
  we
  set $M:=2$ and $L:=1/h$
  (cf.\ (\ref{eq:Vitushkin})).
  \qedtext
\end{proof}

The regret term in (\ref{eq:analytic-from-analytic})
is not quite as good as the regret term $\frac{C}{h}\log^2 N$
that would be obtained if we used the right value $h$ instead of using $H$
(cf.\ (\ref{eq:Banach-analytic-2})),
but the difference is not great.

\subsection*{From classes of type II to classes of type III}

In this subsection we will see how well prediction strategies
designed for type II classes can cope with type III classes.
The difference between the sizes of the classes of different types is huge,
and the leap might lead to losing half of the smoothness
of type III classes.

\begin{lemma}\label{lem:approximation-by-analytic}
  Let $h>0$ and let $F:\bbbr\to\bbbr$ be a non-zero periodic function with period $2\pi$
  whose $k$th derivative ($k\in\{0,1,\ldots\}$) exists
  and is H\"older continuous of order $\alpha\in(0,1]$
  with coefficient $c$.
  Set $s:=k+\alpha$.
  For small enough $\epsilon>0$,
  \begin{equation}\label{eq:approximation-by-analytic}
    \log\AAA_{\epsilon}^{A_h}(F)
    \le
    C h
    \left(
      \frac{12c}{\epsilon}
    \right)^{1/s},
  \end{equation}
  where $C$ is a universal constant.
\end{lemma}
\begin{proof}
  We will emulate the proof of Lemma \ref{lem:analytic-by-analytic}.
  According to Jackson's theorem
  (\cite{natanson:1964}, Theorem 2 in \S IV.3)
  there is a trigonometric polynomial of degree $J$ at a uniform distance
  from $F$ at most
  \begin{equation*}
    \frac{12^{k+1}c(1/J)^{\alpha}}{J^k}
    =
    12^{k+1}cJ^{-s}.
  \end{equation*}
  This distance will not exceed $\epsilon>0$
  if we set
  \begin{equation}\label{eq:J-2}
    J
    :=
    \left\lceil
      \left(
        \frac{12^{k+1}c}{\epsilon}
      \right)^{1/s}
    \right\rceil.
  \end{equation}
  The absolute value of the approximating trigonometric polynomial
  does not exceed
  \begin{equation}\label{eq:max-h}
    \left(
      \left\|
        F
      \right\|_{C(\bbbr)}
      +
      \epsilon
    \right)
    e^{Jh}
  \end{equation}
  (cf.\ (\ref{eq:max}))
  in the strip $\left|\Im z\right|<h$,
  and so we can substitute (\ref{eq:J-2}) into (\ref{eq:max-h})
  to find
  \begin{equation*}
    \log\AAA_{\epsilon}^{A_h}(F)
    \le
    \log
    \left(
      \left\|
        F
      \right\|_{C(\bbbr)}
      +
      \epsilon
    \right)
    +
    (\log e)
    h
    \left\lceil
      \left(
        \frac{12^{k+1}c}{\epsilon}
      \right)^{1/s}
    \right\rceil,
  \end{equation*}
  which for small enough $\epsilon$ gives (\ref{eq:approximation-by-analytic})
  with any $C>12\log e$.
  \qedtext
\end{proof}

\begin{remark*}
  Another way of deriving an estimate for $\AAA_{\epsilon}^{A_h}(F)$
  for a smooth $F$
  would be to combine Kolmogorov's estimate \cite{kolmogorov:1935}
  of the remainder of the Fourier series 
  with the known results about the size of coefficients in Fourier series
  (\cite{\Bary}, \S I.24).
  This would, however, produce a weaker result.
\end{remark*}

Combining Lemma \ref{lem:approximation-by-analytic}
with Theorem \ref{thm:continuous},
we now obtain:

\begin{corollary}\label{cor:smooth-from-analytic}
  Let $F:\bbbr\to\bbbr$ be a periodic period $2\pi$ function
  whose $k$th derivative ($k\ge0$) is H\"older continuous
  of order $\alpha$ with coefficient $c$.
  The strategy for Predictor constructed for the class $A_h$
  guarantees
  \begin{equation}\label{eq:smooth-from-analytic}
    \sum_{n=1}^N
    \left(
      y_n-\mu_n
    \right)^2
    \le
    \sum_{n=1}^N
    \left(
      y_n-F(x_n)
    \right)^2
    +
    C
    h^{\frac{s}{s+2}}
    c^{\frac{2}{s+2}}
    N^{\frac{2}{s+2}}
  \end{equation}
  from some $N$ on,
  where $s:=k+\alpha$
  and $C$ is a universal constant.
\end{corollary}

\begin{proof}
  The proof is similar to that of Corollary \ref{cor:analytic-from-analytic}.
  The regret term in (\ref{eq:continuous-analytic})
  can be bounded above by
  \begin{equation}\label{eq:bound}
    C'_M
    \inf_{\epsilon\in\left(0,1\right]}
    \left(
      L
      h^M
      \left(
        \frac{12c}{\epsilon}
      \right)^{M/s}
      +
      \epsilon N
    \right)
  \end{equation}
  (it is clear that the term $L\log^M\frac{1}{\epsilon}$
  can be ignored).
  Using the upper bound (\ref{eq:approx-min}) for (\ref{eq:bound}),
  we obtain
  \begin{equation*}
    2C'_M
    L^{\frac{s}{s+M}}
    h^{\frac{Ms}{s+M}}
    (12c)^{\frac{M}{s+M}}
    N^{\frac{M}{s+M}}.
  \end{equation*}
  Ignoring $12^{M/(s+M)}\in(1,12)$ and substituting $M:=2$ and $L:=1/h$
  (cf.\ (\ref{eq:Vitushkin})),
  we reduce this to (\ref{eq:smooth-from-analytic}).
  \qedtext
\end{proof}

The growth rate $N^{2/(s+2)}=N^{1/(s/2+1)}$ of the regret term
in (\ref{eq:smooth-from-analytic})
is worse than the rate $N^{1/(s+1)}$
obtained in \S\ref{sec:Sobolev}
(see (\ref{eq:specific}))
for a prediction strategy designed specifically
for functions with H\"older continuous derivatives.
We can say that one loses half of the smoothness of $F$
when using the wrong benchmark class.

\subsection*{Domination between Sobolev-type classes}

We first state a trivial corollary of the definition of real interpolation
in terms of the K-method
(in the form of the ``approximation theorem''
in \cite{adams/fournier:2003}, 5.31--5.32).
For the definition of the K-method,
see, e.g., 
\cite{bergh/lofstrom:1976}, \S3.1,
or \cite{adams/fournier:2003}, 7.8--7.10;
the notation $(X_0,X_1)_{\theta,q}$ below
can be understood to be the abbreviation for $(X_0,X_1)_{\theta,q,K}$.
We will be mostly interested in the case $q=\infty$.

\begin{lemma}\label{lem:K-method}
  Let $(X_0,X_1)$ be an interpolation pair
  and $\theta\in(0,1)$;
  set $X:=(X_0,X_1)_{\theta,\infty}$.
  For each $F\in X$ and each $t>0$
  there exists $F_{t}\in X_1$ such that
  \begin{equation}\label{eq:K-method}
    \begin{cases}
      \left\|
        F - F_{t}
      \right\|_{X_0}
      \le
      2 t^{\theta}
      \left\|
        F
      \right\|_{X}\\
      \left\|
        F_{t}
      \right\|_{X_1}
      \le
      2 t^{\theta-1}
      \left\|
        F
      \right\|_{X}.
    \end{cases}
  \end{equation}
\end{lemma}

\begin{proof}
  Since the function 
  \begin{equation*}
    K(t,F)
    :=
    \inf_{F_0\in X_0,F_1\in X_1: F=F_0+F_1}
    \left(
      \left\|
        F_0
      \right\|_{X_0}
      +
      t
      \left\|
        F_1
      \right\|_{X_1}
    \right)
  \end{equation*}
  (this is a generalization of (\ref{eq:K}))
  is continuous in $t$ (\cite{bergh/lofstrom:1976}, Lemma 3.1.1),
  we have, by the definition of the K-method:
  \begin{multline*}
    \left\|
      F
    \right\|_{X}
    =
    \sup_{t\in(0,\infty)}
    t^{-\theta}
    K(t,F)\\
    =
    \sup_{t\in(0,\infty)}
    \inf
    \left\{
      t^{-\theta}
      \left\|
        F_0
      \right\|_{X_0}
      +
      t^{1-\theta}
      \left\|
        F_1
      \right\|_{X_1}
      \stbig
      F = F_0 + F_1,
      F_0 \in X_0,
      F_1 \in X_1
    \right\}.
  \end{multline*}
  Therefore, for each $t>0$ there is a split $F = F_0 + F_1$ such that
  \begin{equation*}
    t^{-\theta}
    \left\|
      F_0
    \right\|_{X_0}
    +
    t^{1-\theta}
    \left\|
      F_1
    \right\|_{X_1}
    \le
    2
    \left\|
      F
    \right\|_{X},
  \end{equation*}
  which is stronger than the statement of the lemma.
  \qedtext
\end{proof}

By \cite{bergh/lofstrom:1976}, Theorem 6.4.5(1),
\begin{equation}\label{eq:Besov}
  s_0 \ne s_1
  \;\Longrightarrow\;
  \left(
    B^{s_0}_{p,q_0},
    B^{s_1}_{p,q_1}
  \right)_{\theta,r}
  =
  B^{(1-\theta)s_0+\theta s_1}_{p,r},
\end{equation}
and applying this to the H\"older--Zygmund spaces
$\CCC^s(\mathbf{X}):=B^s_{\infty,\infty}(\mathbf{X})$
we obtain the following corollary of Lemma \ref{lem:K-method}.

\begin{corollary}\label{cor:K-method}
  Let $0<s<S<\infty$.
  For each $F\in\CCC^s(\mathbf{X})$ and each $\epsilon>0$
  there exists $F_{\epsilon}\in\CCC^S(\mathbf{X})$ such that
  \begin{equation*}
    \begin{cases}
      \left\|
        F - F_{\epsilon}
      \right\|_{C(\mathbf{X})}
      \le
      C \epsilon^{s}
      \left\|
        F
      \right\|_{\CCC^s(\mathbf{X})}\\
      \left\|
        F_{\epsilon}
      \right\|_{\CCC^S(\mathbf{X})}
      \le
      2 \epsilon^{s-S}
      \left\|
        F
      \right\|_{\CCC^s(\mathbf{X})},
    \end{cases}
  \end{equation*}
  where $C$ is a universal constant.
\end{corollary}
\begin{proof}
  Setting $\theta:=s/S$, we obtain from (\ref{eq:Besov}):
  \begin{equation*}
    \left(
      B^{0}_{\infty,1},
      B^{S}_{\infty,\infty}
    \right)_{s/S,\infty}
    =
    B^{s}_{\infty,\infty}.
  \end{equation*}
  Remember that there is a continuous embedding
  $B^0_{\infty,1}(\mathbf{X})\hookrightarrow C(\mathbf{X})$
  (\cite{edmunds/triebel:1996}, (2.3.3/3)).
  It remains to set $t:=\epsilon^S$ in (\ref{eq:K-method}).
  \qedtext
\end{proof}

Let us apply the last corollary to the case
of the performance bound (\ref{eq:performance-Holder-1}).
That bound
(together with its derivation,
involving the $\CCC^s(\mathbf{X})$ norm)
gives the regret term
of order, approximately,
\begin{equation}\label{eq:s-1}
  \left\|
    F
  \right\|_{\CCC^s(\mathbf{X})}
  N^{1-s/m}
\end{equation}
for the benchmark class $\CCC^s(\mathbf{X})$,
$0<s\le m/2$,
and of order
\begin{equation}\label{eq:S-1}
  \left\|
    F
  \right\|_{\CCC^S(\mathbf{X})}
  N^{1-S/m}
\end{equation}
for the benchmark class $\CCC^S(\mathbf{X})$,
$0<S\le m/2$.
Suppose $s<S$ and let us see
when a prediction strategy ensuring regret term (\ref{eq:S-1})
for $\CCC^S(\mathbf{X})$
automatically ensures regret term (\ref{eq:s-1})
for $\CCC^s(\mathbf{X})$.

Corollary \ref{cor:K-method} guarantees that every prediction strategy
ensuring regret term (\ref{eq:S-1})
for $F\in\CCC^S(\mathbf{X})$
ensures regret term
\begin{multline}\label{eq:regret-1}
  \inf_{\epsilon>0}
  \left(
    \left\|
      F_{\epsilon}
    \right\|_{\CCC^S(\mathbf{X})}
    N^{1-S/m}
    +
    \left\|
      F - F_{\epsilon}
    \right\|_{C(\mathbf{X})}
    N
  \right)\\
  \le
  \inf_{\epsilon>0}
  \left(
    2
    \left\|
      F
    \right\|_{\CCC^s(\mathbf{X})}
    N^{1-S/m}
    \epsilon^{s-S}
    +
    C
    \left\|
      F
    \right\|_{\CCC^s(\mathbf{X})}
    N
    \epsilon^s
  \right)
\end{multline}
for $F\in\CCC^s(\mathbf{X})$.
Using the upper bound (\ref{eq:approx-min}) for (\ref{eq:regret-1}),
we obtain regret
\begin{equation*}
  2
  \left(
    2
    \left\|
      F
    \right\|_{\CCC^s(\mathbf{X})}
    N^{1-S/m}
  \right)^{\frac{s}{S}}
  \left(
    C
    \left\|
      F
    \right\|_{\CCC^s(\mathbf{X})}
    N
  \right)^{\frac{S-s}{S}},
\end{equation*}
which coincides, to within a constant factor, with (\ref{eq:s-1}).

We can see that the case $s\approx m/2$ in (\ref{eq:performance-Holder-1})
dominates all other cases with $s\le m/2$.
The bound for the case $s\approx m/2$ was derived in \cite{\DFV}
using Hilbert-space methods (applicable when $p=2$).
The Banach-space methods developed in \cite{\DFVI}
might eventually turn out to be less important
(but remember that we only considered Besov spaces $B^s_{p,q}$
with $p$ and $q$ set to infinity).

To extend this analysis to the case $s>m/2$,
we will have to compare regret terms of order
\begin{equation}\label{eq:s-2}
  \left\|
    F
  \right\|_{\CCC^s(\mathbf{X})}^{\frac{m}{m+s}}
  N^{\frac{m}{m+s}}
\end{equation}
for the benchmark class $\CCC^s(\mathbf{X})$
and
\begin{equation}\label{eq:S-2}
  \left\|
    F
  \right\|_{\CCC^S(\mathbf{X})}^{\frac{m}{m+S}}
  N^{\frac{m}{m+S}}
\end{equation}
for $\CCC^S(\mathbf{X})$,
where $0<s<S$
(see (\ref{eq:Banach-Sobolev-Besov}) 
with $p$ and $q$ set to $\infty$,
as in the case of (\ref{eq:specific})).
Since our comparison is informal anyway,
we will ignore the $\log\log$ terms in (\ref{eq:Banach-Sobolev-norm}).
Corollary \ref{cor:K-method} and the upper bound (\ref{eq:approx-min})
imply that every prediction strategy
ensuring regret term (\ref{eq:S-2})
for $\CCC^S(\mathbf{X})$
will also ensure regret term
\begin{multline}\label{eq:not-as-good}
  \inf_{\epsilon>0}
  \left(
    \left\|
      F_{\epsilon}
    \right\|_{\CCC^S(\mathbf{X})}^{\frac{m}{m+S}}
    N^{\frac{m}{m+S}}
    +
    \left\|
      F - F_{\epsilon}
    \right\|_{C(\mathbf{X})}
    N
  \right)\\
  \le
  \inf_{\epsilon>0}
  \left(
    2
    \left\|
      F
    \right\|_{\CCC^s(\mathbf{X})}^{\frac{m}{m+S}}
    N^{\frac{m}{m+S}}
    \epsilon^{(s-S)\frac{m}{m+S}}
    +
    C
    \left\|
      F
    \right\|_{\CCC^s(\mathbf{X})}
    N
    \epsilon^s
  \right)\\
  \le
  C'
  \left(
    \left\|
      F
    \right\|_{\CCC^s(\mathbf{X})}^{\frac{m}{m+S}}
    N^{\frac{m}{m+S}}
  \right)^{\frac{s(m+S)}{S(m+s)}}
  \left(
    \left\|
      F
    \right\|_{\CCC^s(\mathbf{X})}
    N
  \right)^{\frac{(S-s)m}{S(m+s)}}\\
  =
  C'
  \left(
    \left\|
      F
    \right\|_{\CCC^s(\mathbf{X})}
    N
  \right)^{\frac{m}{m+s}}
\end{multline}
for $\CCC^s(\mathbf{X})$.
The regret rate obtained is as good as (\ref{eq:s-2}),
to within a constant factor.
Therefore,
as far as our bounds are concerned,
$\CCC^S(\mathbf{X})$ dominates $\CCC^s(\mathbf{X})$.
Unfortunately,
these bounds are known to be loose (see \S\ref{sec:Sobolev}),
at least in the case of low smoothness,
and it remains to be seen whether the domination still holds
for tighter bounds.

\section{Conclusion}
\label{sec:conclusion}

In this paper we have seen the following typical rates of growth
of the regret term:
\renewcommand{\labelenumi}{(\Roman{enumi})}
\begin{enumerate}
\item
  for finite dimensional $\FFF$ (type I of \cite{\KolmogorovTikhomirov}, \S3),
  \begin{equation*}
    O
    \left(
      \log N
    \right);
  \end{equation*}
\item
  for classes $\FFF$ of analytic functions of $m$ variables
  (type II of \cite{\KolmogorovTikhomirov}),
  \begin{equation*}
    O
    \left(
      \log^{m+1} N
    \right);
  \end{equation*}
\item
  for classes $\FFF$ of functions of $m$ variables
  with smoothness indicator $s$
  (type III of \cite{\KolmogorovTikhomirov}),
  \begin{equation*}
    O
    \left(
      N^{\frac{m}{m+s}}
    \right);
  \end{equation*}
\item
  for classes $\FFF$ of Lipschitzian functionals
  on classes of the previous type
  (such $\FFF$ are representative of type IV of \cite{\KolmogorovTikhomirov}),
  a typical rate is
  \begin{equation*}
    O
    \left(
      N / \log^{s/m} N
    \right).
  \end{equation*}
\end{enumerate}
Rates of types I and III have been known in competitive on-line prediction,
whereas types II and IV appear new.
For the first time we can see enough fragments
to get an impression of the big picture.
These are still small fragments and the picture is still vague.
The method of metric entropy, despite its wide applicability,
is not universal and often does not give optimal results.
My goal was to convince my listeners or readers
that the arising questions are interesting ones.

We have also considered, in a very tentative way,
the question of how much one has to pay for using a wrong benchmark class
(\S\ref{sec:super}).
From the available very preliminary results
it appears that using meagre (albeit dense in $C(\mathbf{X})$) benchmark classes
is safer than using rich classes.

These are possible directions of theoretical research:
\begin{itemize}
\item
  Find computationally efficient prediction strategies for benchmark classes
  such as $A_G^K$ and $A_h$ (type II)
  and Besov spaces with $m/(m+s)<1/2$
  (in the notation of (\ref{eq:Banach-Sobolev-Besov})).
\item
  Find uniform in $N$ estimates of metric entropy
  (for applications such as those in \S\S\ref{sec:norm}--\ref{sec:super}).
\item
  Extend this paper's results to discontinuous prediction rules
  (for estimates of metric entropy in this case
  see, e.g., \cite{clements:1963}).
\item
  Perhaps most importantly,
  complement performance guarantees such as those in this paper
  with lower bounds.
  A lower bound corresponding to Proposition \ref{prop:DFVI} with $p=2$
  is proved in \cite{\DFV}, Theorem 4;
  however, the function space $\FFF$ constructed there
  is not compactly embedded in $C(\mathbf{X})$,
  and so not interesting from the point of view of metric entropy.
\end{itemize}

In experimental research,
it would be interesting to find out the ``empirical approachability function''
\begin{equation}\label{eq:empirical-approachability}
  \AAA_{\epsilon}^{\FFF,2}
  \left(
    x_1,y_1,
    \ldots,
    x_N,y_N
  \right)
  :=
  \inf
  \left\{
    \left\|
      F
    \right\|_{\FFF}
    \stbegin
    \frac1N
    \sum_{n=1}^N
    \left(
      y_n - F(x_n)
    \right)^2
    \le
    \epsilon
    \stend
  \right\}
\end{equation}
(cf.\ (\ref{eq:approachability});
the upper index 2 refers to using the quadratic loss function in this definition)
for standard benchmark data sets
\begin{equation}\label{eq:data-set}
  \left(
    x_1,y_1,
    \ldots,
    x_N,y_N
  \right)
  :=
  \left(
    (x_1,y_1),
    \ldots,
    (x_N,y_N)
  \right)
\end{equation}
and standard function classes $\FFF$.
It is clear that (\ref{eq:empirical-approachability}) will be finite for all $\epsilon>0$
if $\FFF$ is dense in $C(\mathbf{X})$ and
\begin{equation*}
  x_{n_1} = x_{n_2}
  \;\Longrightarrow\;
  y_{n_1} = y_{n_2}.
\end{equation*}
If a prediction strategy guarantees a regret term of $f(\left\|F\right\|_{\FFF},N)$
(we will assume that $f$ is a continuous function in its first argument),
in the sense that
\begin{equation*}
  \sum_{n=1}^N
  \left(
    y_n-\mu_n
  \right)^2
  \le
  \sum_{n=1}^N
  \left(
    y_n-F(x_n)
  \right)^2
  +
  f(\left\|F\right\|_{\FFF},N)
\end{equation*}
for all $F\in\FFF$ and all $N=1,2,\ldots$,
the loss of this prediction strategy on the data set (\ref{eq:data-set}) will be at most
\begin{equation*}
  \inf_{\epsilon>0}
  \Bigl(
    f
    \left(
      \AAA_{\epsilon}^{\FFF,2}
      \left(
        x_1,y_1,
        \ldots,
        x_N,y_N
      \right),
      N
    \right)
    +
    \epsilon N
  \Bigr).
\end{equation*}
Knowing typical empirical approachability functions (\ref{eq:empirical-approachability})
for various function classes might suggest function classes
most promising for various practical problems.

A natural next step would be to compare
different benchmark classes on real-world data sets.
This is a task for experimental machine learning;
what learning theory can do is to study the relation of domination
between various \emph{a priori} plausible benchmark classes:
e.g., some of them may turn out to be useless or nearly useless
on purely theoretical grounds.

\subsection*{Acknowledgments}

This paper was written to support my talk at the workshop
``Metric entropy and applications in analysis, learning theory and probability''
(Edinburgh, Scotland, September 2006).
I am grateful to its organizers,
Thomas K\"uhn, Fernando Cobos and W.~D.~Evans,
for inviting me.
This version of the paper is preliminary
and is likely to be revised
as a result of discussions at the workshop.

Nicol\`o Cesa-Bianchi,
G\'abor Lugosi,
Steven Smale
and Alex Smola
supplied the principal components of this paper
with their incisive questions and comments.
Ilia Nouretdinov's help was invaluable.
This work was partially supported by MRC (grant S505/65).

\end{document}